\documentclass[10pt, a4paper]{article}

\usepackage[]{lrec2026} 

\usepackage{booktabs} 
\usepackage{fontawesome5}
\usepackage{multirow}
\usepackage[table,xcdraw]{xcolor}
\usepackage{graphicx}
\usepackage{multirow}
\usepackage{tikz}
\usepackage{enumitem}
\usepackage{amssymb}       
\usepackage{amsmath}       
\usepackage{graphicx}   
\usepackage{caption} 
\usepackage{listings}
\usepackage{subcaption}
\usetikzlibrary{positioning, shapes, calc}
\usepackage{inconsolata}
\usepackage[scaled=0.9]{beramono}  
\usepackage{alltt}  

\newcommand{\figmemes}{\textsc{FigMemes}}
\newcommand{\memotion}{\textsc{Memotion }2}
\newcommand{\metmeme}{\textsc{MET-Meme}}

\newcommand{\aya}{\texttt{Aya}}
\newcommand{\gemma}{\texttt{Gemma}}
\newcommand{\qwen}{\texttt{Qwen}}
\newcommand{\ayas}{\texttt{Aya-8B}}
\newcommand{\ayal}{\texttt{Aya-32B}}
\newcommand{\gemmas}{\texttt{Gemma-4B}}
\newcommand{\gemmam}{\texttt{Gemma-12B}}
\newcommand{\gemmal}{\texttt{Gemma-27B}}
\newcommand{\qwens}{\texttt{Qwen-7B}}
\newcommand{\qwenm}{\texttt{Qwen-32B}}
\newcommand{\qwenl}{\texttt{Qwen-72B}}

\newcolumntype{g}{>{\columncolor{gray!20}}r}

\definecolor{skyblue}{RGB}{135,206,235}  

\title{I Came, I Saw, I Explained:\\
Benchmarking Multimodal LLMs on Figurative Meaning in Memes}

\usepackage{fontawesome5}
\newcommand{\mainlp}{\faMountain}
\newcommand{\mcml}{\faRobot}
\newcommand{\nrc}{\faCanadianMapleLeaf}

\name{Shijia Zhou, Barbara Plank, Diego Frassinelli}

\name{Shijia Zhou\kern0pt\textsuperscript{\mainlp\kern0pt\mcml}
\quad Saif M. Mohammad\kern0pt\textsuperscript{\nrc\kern0pt} 
\quad Barbara Plank\kern0pt\textsuperscript{\mainlp\kern0pt\mcml}
\quad Diego Frassinelli\kern0pt\textsuperscript{\mainlp\kern0pt} 
}

\address{
\textsuperscript{\mainlp\kern-0pt} MaiNLP, Center for Information and Language Processing, LMU Munich, Germany \\
\textsuperscript{\mcml\kern-0pt} Munich Center for Machine Learning (MCML), Munich, Germany \\
\textsuperscript{\nrc\kern0pt} National Research Council Canada, Ottawa, Canada \\
\texttt{\{zhou.shijia, b.plank, diego.frassinelli\}@lmu.de}
\\
[0.5ex]
}

\abstract{
Internet memes represent a popular form of multimodal online communication and often use figurative elements to convey layered meaning through the combination of text and images. 
However, it remains largely unclear how multimodal large language models (MLLMs) combine and interpret visual and textual information to identify figurative meaning in memes.
To address this gap, we evaluate eight state-of-the-art generative MLLMs across three datasets on their ability to detect and explain six types of figurative meaning. 
In addition, we conduct a human evaluation of the explanations generated by these MLLMs, assessing whether the provided reasoning supports the predicted label and whether it remains faithful to the original meme content.
Our findings indicate that all models exhibit a strong bias to associate a meme with figurative meaning, even when no such meaning is present.
Qualitative analysis further shows that correct predictions are not always accompanied by faithful explanations.
 \\ \newline \Keywords{figurative meaning, internet memes, multimodal large language model} \\ [1ex] }

\begin{document}

\maketitleabstract

\section{Introduction}
\label{sec:introduction}

\begin{figure}[t]
    \centering
    \includegraphics[width=0.98\linewidth]{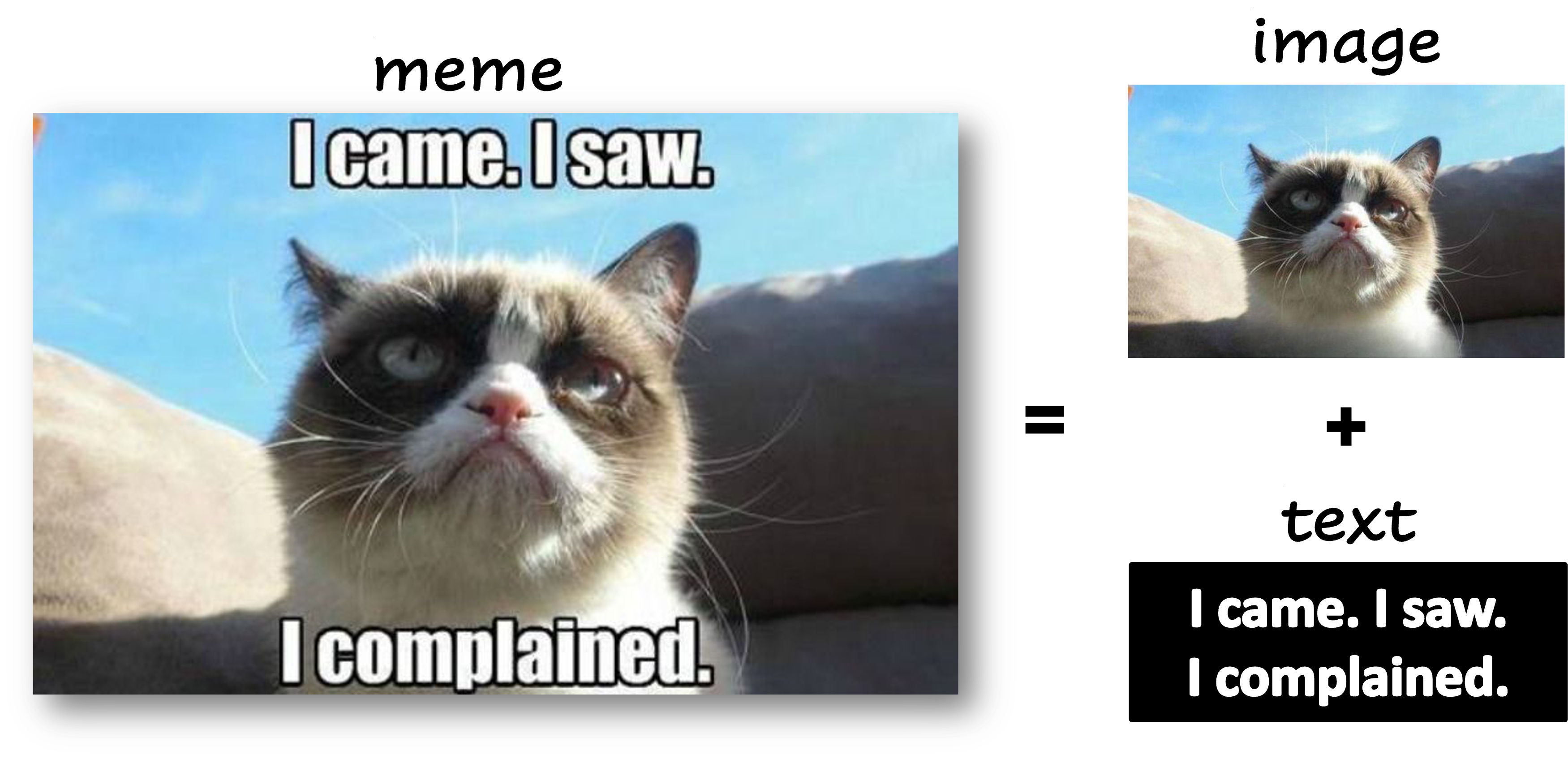}
    \caption{A meme from \metmeme~\citep{xu2022METMeme} playfully adapts Caesar’s famous quote ``I came, I saw, I conquered''.
    In our work, we investigate the respective contributions of image and text to models' prediction of figurative meaning in memes.
    }
    \label{fig:figure1}
\end{figure}

On social media such as Reddit and X (Twitter), people actively use memes, which combine visual and textual components, to express a wide range of opinions.
Memes are posted in debate on global issues, such as climate change~\citep{zhou2025What}, as well as in playful reflections on daily life’s small triumphs and troubles~\citep{hwang2023MemeCap}.
Many of these memes rely on multimodal figurative expressions, to convey meanings in creative, often non-literal ways.
For example, Figure~\ref{fig:figure1} shows a meme from \metmeme~\citep{xu2022METMeme}, which humorously conveys a witty or self-deprecating attitude.
It parodies Julius Caesar’s famous quote ``I came, I saw, I conquered'' by keeping the initial ``I came, I saw'', which sets up the expectation of a powerful declaration, but replaces the last part with ``I complained''.
Together with the accompanying grumpy cat, this subverts the expectation and creates a humorous contrast.

The presence of figurative meaning in memes represents a challenging aspect of multimodal communication,\footnote{In this paper, we use the term \textit{figurative meaning} instead of the more commonly used \textit{figurative language} to emphasize the multimodal nature of the phenomenon we are analyzing.} and has become a phenomenon of growing interest in NLP.
~\citet{liu2022FigMemes} first introduced \figmemes, a dataset for classifying six types of figurative expressions in politically opinionated memes, covering diverse themes and visual styles. Metaphor processing attracts attention in meme interpretation tasks~\citep{xu2022METMeme, hwang2023MemeCap, xu2024Exploring}, and several studies have linked offensive memes with sarcastic expressions \citep{sharma-etal-2020-semeval, pramanick2022Multimodal, kumari2024CMOffMeme}.

However, except for \figmemes{}~\cite{liu2022FigMemes},  figurative meaning in memes remains understudied.
Little work has investigated to what extent generative multimodal large language models (MLLMs) can detect figurative meaning in memes, and how this ability varies across different types, such as allusion, irony, and metaphor.
There is also a lack of systematic investigation into how controlled manipulations of information from different modalities affect figurative meaning predictions on memes.

Recently, natural language explanations have become an important additional signal for studying how a model processes an input, e.g., for interpreting figurative meaning~\cite{saakyan-etal-2025-understanding}.
However, there is a lack of well-defined criteria for evaluating the quality of model-generated explanations of figurative meaning.

To address these gaps, we define a combined \textit{detection and explanation} task for memes, requiring models not only to identify one or more types of figurative meaning, but also provide a corresponding explanation that justifies their predictions. Our contributions are: 

\begin{itemize}[topsep=1pt, partopsep=1pt, itemsep=2pt, parsep=1pt]
    \item We introduce a detection and explanation task for figurative meaning in memes, and evaluate the performance of eight models on both \textit{binary} and \textit{multi-label} classification tasks.
    \item We systematically investigate the impact and contribution of information from \textit{different modalities} on model behavior through controlled input manipulations, and two prompting setups.
    \item We conduct a human study to assess how coherently model-provided explanations are perceived for their respective predicted labels, and how well they are grounded in the meme's visual and textual components.
\end{itemize}

Our benchmarking of eight state-of-the-art MLLMs reveals three core findings regarding their interpretation of meme figurativeness.
First, we identify a pervasive bias where models consistently over-assign figurative meanings to memes that are intended to be literal.
Second, model performance and modality reliance are highly type-specific: while irony and sarcasm are heavily visually grounded, metaphor detection uniquely requires a complex interplay between textual and visual cues.
Finally, our human evaluation uncovers a significant \emph{faithfulness gap}: even when models predict correct labels, their explanations often suffer from visual hallucinations, over-interpretation, or an inability to grasp the underlying socio-psychological nuances of human behavior.
Our code and evaluation are publicly available at \url{https://github.com/mainlp/Figurative-Meaning-in-Memes}.

\section{Related Work}
\label{sec:related_work}

\paragraph{Figurative Language in NLP}

Figurative language is characterized by the use of words in a non-literal manner to create abstract meanings that go beyond surface-level interpretation~\citep{lakoff2008metaphors}. This may facilitate and intensify
indirect communication and rhetorical effects, serving persuasive and humorous discourse, or audience engagement~\citep{fussell2014figurative, burgers2016Figurativea}.

In NLP, figurative language has traditionally been studied through a text-centric lens~\citep{chakrabarty-etal-2022-rocket, stowe-etal-2022-impli, lai-etal-2023-multilingual}.
With the advance of MLLMs, research has shifted toward systematic benchmarking of their ability to process these non-literal expressions.
This includes broad investigations into the models' general reasoning capabilities and linguistic nuances~\citep{liu-etal-2022-testing, jang-etal-2023-figurative, yerukola-etal-2024-pope, jang-frassinelli-2024-generalizable}, as well as their performance in specialized interactive contexts such as dialogue systems~\citep{jhamtani-etal-2021-investigating}.

Despite the emphasis on text, figurative meaning is inherently a multimodal phenomenon~\citep{chakrabarty-etal-2022-flute, akula2023metaclue, kulkarni-etal-2024-report}. With the increasing availability of multimodal generative models, researchers started investigating non-literal meaning across combined modalities, with work on humor understanding in comics~\citep{hessel-etal-2023-androids}, on generating visualizations of textual metaphors~\citep{zhang-etal-2024-gome}, and on captioning visual figurative meanings~\cite{saakyan-etal-2025-understanding}.

\paragraph{Multimodal Meme Understanding}
Internet memes are a distinctive form of multimodal communication, where visual components shape the interpretation of text through context, priming, or template-based expectations~\citep{shifman2013memes, nissenbaum2017Internet, wiggins2019discursive}.
The NLP and vision-language communities have shown a growing interest in memes, with research highlighting how visual templates provide a contextual framework that shapes the interpretation and structure of the associated text~\citep{zhou-etal-2024-social, bates-etal-2025-template}.
Various tasks have been explored, including sentiment analysis~\citep{hossain-etal-2022-memosen}, hateful or harmful meme detection~\citep{kiela2020hateful, cao-etal-2022-prompting, liu2025MIND}, emotion classification~\citep{sharma-etal-2020-semeval}, caption generation~\citep{hwang2023MemeCap}, style generalization~\citep{nandy2024YesBut}, and VQA tasks~\citep{nguyen2025MemeQA}.

Although memes often contain a wide range of figurative meaning, few studies examine how well models handle them.
\citet{liu2022FigMemes} laid groundwork for multimodal pre-LLM figurative meaning analysis and introduced \figmemes, a collection of politically-opinionated memes. 

To the best of our knowledge, our study is the first to systematically  evaluate multimodal models in zero-shot settings by explicitly separating visual and textual inputs, enabling us to precisely assess the individual contribution of each modality.
Moreover, we move beyond label accuracy to evaluate the quality of model-generated explanations, providing a more comprehensive benchmark for the interpretation of figurative meaning.

\begin{table*}[ht]
\centering
\resizebox{0.98\textwidth}{!}{
\begin{tabular}{l | c c c c c c c c | c c c c c | c c c}
\toprule
 & \multicolumn{8}{c|}{\textbf{\figmemes}} & \multicolumn{5}{c|}{\textbf{\memotion}} & \multicolumn{3}{c}{\textbf{\metmeme}} \\
\textbf{Type} & \underline{None} & Allus. & Exag. & Irony & Anthrop. & Met & Contr. & \cellcolor{gray!20}\textit{Total} &
\underline{NS} & LS & VS & ES & \cellcolor{gray!20}\textit{Total} &
\underline{Lit} & Met & \cellcolor{gray!20}\textit{Total} \\
\textbf{Count} & 495 & 265 & 265 & 320 & 131 & 286 & 171 & \cellcolor{gray!20}1372 &
804 & 388 & 246 & 62 & \cellcolor{gray!20}1500 &
1054 & 1054 & \cellcolor{gray!20}2108 \\
\bottomrule
\end{tabular}
}
\caption{Statistics of our evaluation data. Underlined \underline{types} indicate negative cases (e.g.\ non figurative) in the binary classification task of each dataset.
\textit{Total} shows the number of evaluated memes in each dataset; in \figmemes, a meme can span multiple types of figurative meaning.
\figmemes: Allus. = Allusion, Exag. = Exaggeration, Anthrop. = Anthropomorphism, Met. = Metaphor, Contr. = Contrast. \memotion: NS, LS, VS, ES indicate increasing levels of sarcasm (Not, Little, Very, Extremely). \metmeme: Lit = Literal, Met = Metaphoric.}
\label{tab:data_statistics}
\end{table*}

\section{Experimental Setup} \label{sec:experiment_setup}

\subsection{Datasets}

We analyze three publicly available meme datasets concerning figurative expressions: \figmemes{}~\citep{liu2022FigMemes}, \memotion{}~\citep{ramamoorthy2022memotion}, and \metmeme{}~\citep{xu2022METMeme}. To ensure modality comparability, only memes with embedded text are evaluated.

\begin{itemize}[leftmargin=1.2em,itemsep=0pt,topsep=0pt]
    \item \textbf{\figmemes} is multi-label and the first and only dataset which annotates six types of figurative language in memes: 
    \texttt{allusion}, 
    \texttt{exaggeration/hyperbole},
    \texttt{irony/sarcasm},
    \texttt{anthropomorphism/zoomorphism},
    \texttt{metaphor/\allowbreak simile} and 
    \texttt{contrast}.
    They collect memes from 4chan /pol/ (the politically incorrect) board,\footnote{\url{http://boards.4chan.org/pol/}} a platform known for its high prevalence of hateful and offensive material.

    \item \textbf{\memotion} annotates memes from Google Images with multiple sentiment and emotion labels, including funny, sarcasm, offensive, and motivational. 
    Here we focus on \texttt{sarcasm} only as it is not only associated with the emotion of contempt, but also characterized by indirectness and irony, which are central to figurative expressions.

    \item \textbf{\metmeme} includes memes from Twitter, Weibo, Google, and Baidu images. It focuses on metaphor occurrences in memes, but also contains manual annotations of sentiment categories, intentions, and offensiveness degrees. While the dataset contains both Chinese and English data, in this work, we focus only on the English memes, and specifically on the binary labels indicating whether a meme is \texttt{metaphorical} or not.
\end{itemize}

\paragraph{Label Distribution}

For evaluation, we use the \figmemes{} test set, covering six figurative types in political memes, and the \memotion{} validation set for sarcasm analysis.\footnote{In \memotion{}~\citep{sharma-etal-2020-semeval, pramanick2022Multimodal}, \textit{little}, \textit{very}, and \textit{extremely sarcastic} are considered positive cases; \textit{not sarcastic} is considered negative. The test set is not publicly available. }
For \metmeme{}, we take all metaphoric items and randomly sample an equal number of literal items.
This selection provides a diverse range of non-literal meanings across different meme domains.
The statistics of our evaluation data are shown in Table~\ref{tab:data_statistics}.

All three datasets were originally annotated by multiple annotators. We rely on these labels which are summarized below. \figmemes{} takes the majority vote of three annotators, who are also the authors of the paper; \memotion{} takes the majority vote of three workers on Amazon Mechanical Turk; and each meme in \metmeme{} is annotated by at least three graduate students and two research assistants with relevant professional knowledge.\footnote{\figmemes{} labels achieved a Fleiss’ $\kappa$ of 0.42, whereas equivalent agreement metrics for metaphor and sarcasm are not reported for the other datasets.}

\paragraph{Preprocessing} \label{para:preprocessing}

In previous work~\citep{liu2022FigMemes, hwang2023MemeCap}, the whole meme -- with its embedded text -- is usually treated as visual input.
Here, to explore the contributions of different modalities to model processing, we create an \textit{image}-only input condition by removing textual information embedded in the original meme, as illustrated in Figure~\ref{fig:figure1}.
We apply PaddleOCR~\citep{cui2025PaddleOCR} to detect text and mask it, and then use LaMa~\citep{suvorov2021Resolutionrobust} for inpainting.
The quality of the inpainted images is manually verified and found to be high (further details about this in Appendix~\ref{sec:image_preprocessing}).

\subsection{Task Setup}
\label{subsec:task_setup}

We explore the extent to which i) MLLMs uncover non-literal meanings in memes and ii) provide faithful explanations of predictions.

\paragraph{Task and Prompt}
\label{para:task_and_prompt}

Unlike prior work~\citep{liu2022FigMemes, pramanick2022Multimodal, xu2022METMeme}, to the best of our knowledge, we are the first to setup \textit{detection and explanation} tasks: 
models must not only determine whether an input contains one or more types of figurative meanings, but also provide a corresponding explanation.
For detection, we follow two different strategies according to the information available in the different resources. For \textit{binary} classification, we detect \texttt{sarcasm} in \memotion{}, \texttt{metaphor} in \metmeme{}, and each type of figurative meaning independently in \figmemes{}.
Additionally, since each meme in \figmemes{} can be associated with multiple types of figurative meaning, we define a \textit{multi-label} classification task exclusive to \figmemes{}.

We focus on a zero-shot evaluation and design two prompting strategies to validate the results. In the first strategy, we simulate the annotation style of the source datasets: we provide definitions of each figurative type, following \figmemes{}~\citep{liu2022FigMemes}, and prompt the model to return the binary label for specific type of figurative meaning, and its explanation to justify the label:

\begin{center}
\begin{tikzpicture}
\node[
    draw,
    rectangle,
    rounded corners,
    fill=gray!10,
    inner sep=8pt,
    text width=0.9\linewidth,
    font=\small,
    align=left
] {
\ttfamily
System:\\
You are an expert linguistic annotator.\\[0.2\baselineskip]

User:\\
You are asked to annotate whether the input\\
\ \ \ \ contains a \{FIGURATIVE TYPE\} expression\\
\ \ \ \ or not.\\[0.2\baselineskip]

Definition:\\
\ \ \ \ \{FIGURATIVE TYPE\}: \{DEFINITION\}\\[0.2\baselineskip]

\{INPUT\}\\[0.2\baselineskip]

Please explain why you are assigning this\\
\ \ \ \ label. Your explanation should clearly\\
\ \ \ \ justify your choice and reference the\\
\ \ \ \ relevant visual and/or textual compon-\\
\ \ \ \ ents in the input.\\[0.2\baselineskip]

Return your answer ONLY as a JSON object\\
\ \ \ \ in this exact format:\\[0.2\baselineskip]

\{\\
\ \ "label": \\
\ \ \ \ \ \ \{FIGURATIVE TYPE\}: "Yes" or "No",\\
\ \ "explanation": \{generated\_explanation\}\\
\}\\[0.2\baselineskip]

ONLY return a valid JSON object in the \\
\ \ \ \ exact format above.
};
\end{tikzpicture}
\end{center}

For multi-label classification, we incorporate the definitions of all six figurative types into a unified prompt, enabling the model to perform simultaneous inference for every category in a single query.

In the second strategy, we prompt the model to provide a continuous, probability-like value between 0 and 1 to indicate the degree to which the input expresses a specific type of figurative meaning. Except for replacing the ``\texttt{label}'' field with a continuous ``\texttt{score}'', the prompt template is identical to the first setting.

\paragraph{Input Modality}
To investigate how information from each modality influences model behavior, we evaluate three input conditions: the original \textit{meme}, as well as two ablation settings: \textit{text}-only, which inputs only the embedded text from the meme, and \textit{image}-only, which removes the embedded text while retaining the visual content (see Section~\ref{para:preprocessing}).
To minimize potential modality bias arising from prompt phrasing, we deliberately exclude the term ``meme'' from the prompt template.

\begin{table}[t]
\centering
\resizebox{0.85\linewidth}{!}{
\begin{tabular}{l r r r r r}
\toprule
\textbf{Labels per Meme} & 0 & 1 & 2 & 3 & 4 \\
\textbf{Meme Count} & 495 & 715 & 277 & 51 & 4 \\
\bottomrule
\end{tabular}
}
\caption{Count of labels per meme in \figmemes.
On average, each meme has 0.93 labels. 0 means no figurativeness is present.
}
\label{tab:figmemes_labels}
\end{table}

\subsection{Evaluation Setup} \label{subsec:evaluation_setup}

Models are required to generate both labels and corresponding explanations.
We evaluate model performance along three aspects:

\begin{enumerate}[topsep=1pt, partopsep=1pt, itemsep=2pt, parsep=1pt]
    \item model label(s) vs.\@ gold label(s);
    \item model label(s) vs.\@ model explanation; 
    \item model explanation vs.\@ original meme content.
\end{enumerate}
The first aspect is assessed automatically, while the latter two are evaluated via human assessment.

\paragraph{Automatic Evaluation}

For binary classification, we report F1 for each figurative type in the text. 

For multi-label classification on \figmemes, we adopt micro-averaging because the gold annotations in \figmemes{} are highly sparse, with most memes containing only a single type of figurative meaning. The count of labels across memes is reported in Table~\ref{tab:figmemes_labels}.
In this setting, micro metrics provide a more stable and reliable estimate of overall performance. We report micro-accuracy, micro-precision, micro-recall and micro-F1. Macro scores are provided in Appendix~\ref{subsec:macro_score}.

All results are averaged over 5 runs; standard deviations are reported in Appendix~\ref{subsec:result_with_std}.

\paragraph{Human Evaluation} 

To conduct the human evaluation, we randomly sample 90 memes, 30 from each dataset; for \figmemes{}, we focus on memes containing a single type of figurative meaning.
We evaluate explanations along two aspects with seven criteria. 
First, we focus on how well the explanation generated by the model supports its generated label (i.e. evaluating model explanation vs.\ label):

\vspace{0.5em}
{
\centering
\begin{tikzpicture}
\node[
    draw,
    rectangle,
    rounded corners,
    fill=cyan!10,
    inner sep=5pt,
    text width=0.9\linewidth,
    font=\small
] {
\begin{enumerate}[label=\arabic*.]
    \item \textbf{Relevance:} The explanation highlights the key features of the label and shows why it applies.
    \item \textbf{Consistency:} The explanation aligns with the label and contains no internal contradictions.
\end{enumerate}
};
\end{tikzpicture}
\par
}
\vspace{0.5em}

Second, we evaluate how well the explanation generated by the model reflects the content of the original meme.
These criteria are taken from \citet{hwang2023MemeCap}:

\vspace{0.5em}
{
\centering
\begin{tikzpicture}
\node[
    draw,
    rectangle,
    rounded corners,
    fill=cyan!10,
    inner sep=5pt,
    text width=0.9\linewidth,
    font=\small
] {
\begin{enumerate}[label=\arabic*., start=3]
    \item \textbf{Correctness:} 
    The explanation accurately conveys the meaning intended by the person who posted the meme.
    \item \textbf{Appropriate Length:} The explanation length is appropriate for conveying the meaning (i.e.\ it is not too verbose).
    \item \textbf{Visual Completeness:} The explanation describes all the important components in the image.
    \item \textbf{Textual Completeness:} The explanation describes all the important components in the text embedded in the meme.
    \item \textbf{Faithfulness:} All components of the explanation are supported by either the visual or text content (i.e. there are no made-up components).
\end{enumerate}
};
\end{tikzpicture}
\par
}
\vspace{0.5em}

The first author of this paper and a bachelor student majoring in computational linguistics hired as student assistant rated the model explanations on a 5-point Likert scale, ranging from \textit{strongly disagree} (score 0) to \textit{strongly agree} (score 4).
After annotating the first 15 memes, the two annotators discussed their evaluations to align their understanding and ensure consistency in subsequent ratings.
The Cohen's $\kappa$ between the two annotators is 0.79 across all evaluated items, indicating substantial agreement. We report the averaged score of two annotators in Section~\ref{sec:results}.

\subsection{Models}

We evaluate eight models from three families of state-of-the-art MLLMs of different sizes: Aya-Vision
(\aya, \citealt{ustun-etal-2024-aya}; 8B and 32B), Gemma 3
(\gemma, \citealt{team2025Gemma}; 4B, 12B, and 27B), and Qwen2.5-VL
(\qwen, \citealt{bai2025Qwen25VLa}; 7B, 32B, and 72B).
All models are evaluated in zero-shot settings with a unified prompt template (as shown in Section~\ref{para:task_and_prompt}) across all model sizes. 
Implementation details are provided in Appendix~\ref{subsec:model_setup}.

\begin{table*}[t]
\centering
\resizebox{0.98\linewidth}{!}{
\begin{tabular}{lll|*{6}{>{\raggedleft\arraybackslash}p{1.5cm}}|c|c|>{\raggedleft\arraybackslash}p{1.5cm}}
\toprule
 &  &  & \multicolumn{6}{c|}{\textbf{\figmemes}} & \multicolumn{1}{r|}{\textbf{\memotion}} & \multicolumn{1}{r|}{\textbf{\metmeme}} & \multicolumn{1}{r}{} \\
\multirow{-2}{*}{\textbf{Model}} & 
\multirow{-2}{*}{\textbf{Size}} & 
\multirow{-2}{*}{\textbf{Input\phantom{ }\phantom{ }\phantom{ }\phantom{ }\phantom{ }\phantom{ }}} & 
\multicolumn{1}{c}{\textbf{Allusion}} & 
\multicolumn{1}{c}{\textbf{Exag.}} & 
\multicolumn{1}{c}{\textbf{Irony}} & 
\multicolumn{1}{c}{\textbf{Anthrop.}} & 
\multicolumn{1}{c}{\textbf{Metaphor}} & 
\multicolumn{1}{c|}{\textbf{Contrast}} & 
\multicolumn{1}{c|}{\textbf{Sarcasm}} & 
\multicolumn{1}{c|}{\textbf{Metaphor}} & 
\multicolumn{1}{c}{\multirow{-2}{*}{\textit{\textbf{Average}}}} \\ 
\midrule
\multicolumn{3}{c|}{\textit{\textbf{Baseline}}} & \cellcolor[HTML]{F6F9FE}52.32 & \cellcolor[HTML]{D8E6FC}44.00 & \cellcolor[HTML]{EDF3FD}49.77 & \cellcolor[HTML]{D0E1FC}41.76 & \cellcolor[HTML]{DBE8FC}44.87 & \cellcolor[HTML]{FFFCF3}56.91 & -- & -- & -- \\\midrule
 &  & Meme & \cellcolor[HTML]{FEE9AA}\underline{70.04} & \cellcolor[HTML]{FFFAEA}58.63 & \cellcolor[HTML]{FFFDF8}56.06 & \cellcolor[HTML]{F5F8FE}51.94 & \cellcolor[HTML]{F0F5FE}50.62 & \cellcolor[HTML]{FDDD7F}\textbf{77.95} & \cellcolor[HTML]{FFF2CF}
\underline{63.48} & \cellcolor[HTML]{FEEEBD}66.66 & \cellcolor[HTML]{FFF5D7}61.92 \\
 &  &  \multicolumn{1}{r|}{\textit{-Text}\phantom{ }\phantom{ }\phantom{ }}  & \cellcolor[HTML]{FFF0C4}65.44 & \cellcolor[HTML]{FFFAEA}58.52 & \cellcolor[HTML]{6199F5}10.53 & \cellcolor[HTML]{E8F0FD}48.39 & \cellcolor[HTML]{BBD3FB}35.91 & \cellcolor[HTML]{FFF8E2}59.99 & \cellcolor[HTML]{EDF3FD}49.64 & \cellcolor[HTML]{FDFDFE}54.16 & \cellcolor[HTML]{E6EFFD}47.82 \\
 & \multirow{-3}{*}{\textbf{8B}} &  \multicolumn{1}{r|}{\textit{-Image}} & \cellcolor[HTML]{B7D1FA}34.78 & \cellcolor[HTML]{A7C6F9}30.16 & \cellcolor[HTML]{FFFCF2}57.07 & \cellcolor[HTML]{5994F5}8.47 & \cellcolor[HTML]{CADDFB}40.09 & \cellcolor[HTML]{8FB7F8}23.46 & \cellcolor[HTML]{FFF3D1}63.01 & \cellcolor[HTML]{FFFCF1}57.28 & \cellcolor[HTML]{C8DBFB}39.29 \\
 \cline{2-12}
 &  & Meme & \cellcolor[HTML]{FEEDBC}66.87 & \cellcolor[HTML]{FFF6DC}61.01 & \cellcolor[HTML]{FFF6DB}61.31 & \cellcolor[HTML]{FFFFFF}54.67 & \cellcolor[HTML]{FFFFFE}54.97 & \cellcolor[HTML]{FEE8A6}70.85 & \cellcolor[HTML]{FFF3CF}63.37 & \cellcolor[HTML]{FEEEBC}66.77 & \cellcolor[HTML]{FFF4D4}62.48 \\
 &  &  \multicolumn{1}{r|}{\textit{-Text}\phantom{ }\phantom{ }\phantom{ }}  & \cellcolor[HTML]{FEEDB9}67.48 & \cellcolor[HTML]{FFF8E5}59.45 & \cellcolor[HTML]{D2E2FC}42.27 & \cellcolor[HTML]{F4F8FE}51.70 & \cellcolor[HTML]{C8DBFB}39.48 & \cellcolor[HTML]{FFF7E1}60.13 & \cellcolor[HTML]{E4EDFD}47.22 & \cellcolor[HTML]{FFFDF8}56.03 & \cellcolor[HTML]{F8FBFE}52.97 \\
\multirow{-6}{*}{\textbf{Aya}}  & \multirow{-3}{*}{\textbf{32B}} &  \multicolumn{1}{r|}{\textit{-Image}} & \cellcolor[HTML]{EAF1FD}48.99 & \cellcolor[HTML]{C2D7FB}37.66 & \cellcolor[HTML]{FFF9E7}59.09 & \cellcolor[HTML]{5C96F5}9.30  & \cellcolor[HTML]{C2D8FB}37.85 & \cellcolor[HTML]{FFF2CE}63.53 & \cellcolor[HTML]{FFF8E3}59.82 & \cellcolor[HTML]{EEF4FE}50.11 & \cellcolor[HTML]{DFEAFD}45.79 \\
\midrule\midrule
 &  & Meme & \cellcolor[HTML]{FEE9AA}\textbf{70.17} & \cellcolor[HTML]{FCFDFE}53.94 & \cellcolor[HTML]{FFF4D3}62.70 & \cellcolor[HTML]{F3F7FE}51.37 & \cellcolor[HTML]{F9FBFE}53.22 & \cellcolor[HTML]{FFFCF4}56.71 & \cellcolor[HTML]{FFF2CF}63.47 & \cellcolor[HTML]{FEEDBC}66.91 & \cellcolor[HTML]{FFF8E3}59.81 \\
 &  &  \multicolumn{1}{r|}{\textit{-Text}\phantom{ }\phantom{ }\phantom{ }}  & \cellcolor[HTML]{FFFAEC}58.13 & \cellcolor[HTML]{FFFAEB}58.36 & \cellcolor[HTML]{ABC8FA}31.19 & \cellcolor[HTML]{F5F9FE}52.06 & \cellcolor[HTML]{FCFDFE}54.05 & \cellcolor[HTML]{FFF5D8}61.84 & \cellcolor[HTML]{FFF5D7}62.05 & \cellcolor[HTML]{FFF8E2}60.02 & \cellcolor[HTML]{FFFFFF}54.71 \\
 & \multirow{-3}{*}{\textbf{4B}} &  \multicolumn{1}{r|}{\textit{-Image}} & \cellcolor[HTML]{B2CDFA}33.29 & \cellcolor[HTML]{D6E5FC}43.46 & \cellcolor[HTML]{FFFAEB}58.33 & \cellcolor[HTML]{4989F4}3.86 & \cellcolor[HTML]{E4EDFD}47.19 & \cellcolor[HTML]{FFF6DB}61.20 & \cellcolor[HTML]{FFF3D1}63.08 & \cellcolor[HTML]{FFF6DA}61.42 & \cellcolor[HTML]{E1ECFD}46.48 \\
 \cline{2-12}
 &  & Meme & \cellcolor[HTML]{FFF7E0}60.27 & \cellcolor[HTML]{FFFDF8}56.09 & \cellcolor[HTML]{FFF3D1}63.01 & \cellcolor[HTML]{F9FBFE}53.00 & \cellcolor[HTML]{D4E3FC}42.87 & \cellcolor[HTML]{FFFBF0}57.48 & \cellcolor[HTML]{FFF2CE}\textbf{63.54} & \cellcolor[HTML]{FEECB7}67.81 & \cellcolor[HTML]{FFFBED}58.01 \\
 &  &  \multicolumn{1}{r|}{\textit{-Text}\phantom{ }\phantom{ }\phantom{ }}  & \cellcolor[HTML]{FFF2CB}64.06 & \cellcolor[HTML]{FFFCF3}56.96 & \cellcolor[HTML]{DEE9FD}45.54 & \cellcolor[HTML]{FCFDFE}53.94 & \cellcolor[HTML]{ADCAFA}31.87 & \cellcolor[HTML]{FFF8E5}59.48 & \cellcolor[HTML]{FFFAEB}58.45 & \cellcolor[HTML]{F4F8FE}51.84 & \cellcolor[HTML]{F8FAFE}52.77 \\
 & \multirow{-3}{*}{\textbf{12B}} &  \multicolumn{1}{r|}{\textit{-Image}} & \cellcolor[HTML]{FAFCFE}53.45 & \cellcolor[HTML]{EDF3FD}49.79 & \cellcolor[HTML]{FFFBED}57.97 & \cellcolor[HTML]{649BF5}11.34 & \cellcolor[HTML]{BCD3FB}35.95 & \cellcolor[HTML]{FEEEBE}66.58 & \cellcolor[HTML]{FFF5D8}61.78 & \cellcolor[HTML]{FFFFFF}54.80 & \cellcolor[HTML]{EAF1FD}48.96 \\
 \cline{2-12}
 &  & Meme & \cellcolor[HTML]{FFF6DB}61.32 & \cellcolor[HTML]{FFF4D5}\underline{62.32} & \cellcolor[HTML]{FFF4D4}62.51 & \cellcolor[HTML]{F9FBFE}52.99 & \cellcolor[HTML]{FFF2CE}\underline{63.64} & \cellcolor[HTML]{FFF3D2}62.82 & \cellcolor[HTML]{FFF3CF}63.38 & \cellcolor[HTML]{FEEBB4}68.21 & \cellcolor[HTML]{FFF4D6}62.15 \\
 &  &  \multicolumn{1}{r|}{\textit{-Text}\phantom{ }\phantom{ }\phantom{ }} & \cellcolor[HTML]{FFFAE9}58.66 & \cellcolor[HTML]{FFF6DD}60.87 & \cellcolor[HTML]{D4E3FC}42.68 & \cellcolor[HTML]{FBFDFE}53.81 & \cellcolor[HTML]{F8FAFE}52.77 & \cellcolor[HTML]{FFF5D9}61.65 & \cellcolor[HTML]{FFF7DF}60.46 & \cellcolor[HTML]{FCFDFE}53.99 & \cellcolor[HTML]{FFFEFA}55.61 \\
\multirow{-9}{*}{\textbf{Gemma}} & \multirow{-3}{*}{\textbf{27B}} &  \multicolumn{1}{r|}{\textit{-Image}} & \cellcolor[HTML]{FFFFFF}54.66 & \cellcolor[HTML]{D0E0FC}41.53 & \cellcolor[HTML]{FFF7DE}60.68 & \cellcolor[HTML]{6CA0F6}13.78 & \cellcolor[HTML]{E4EDFD}47.18 & \cellcolor[HTML]{FEEEBC}66.80 & \cellcolor[HTML]{FFF4D3}62.77 & \cellcolor[HTML]{FFFBEF}57.65 & \cellcolor[HTML]{F0F5FE}50.63 \\
\midrule\midrule
 &  & Meme & \cellcolor[HTML]{FFF8E2}59.92 & \cellcolor[HTML]{F8FBFE}52.96 & \cellcolor[HTML]{F7FAFE}52.58 & \cellcolor[HTML]{E9F0FD}48.52 & \cellcolor[HTML]{5C95F5}9.10 & \cellcolor[HTML]{FEE18B}75.62 & \cellcolor[HTML]{FFF3D0}63.22 & \cellcolor[HTML]{FEE8A6}\textbf{70.90} & \cellcolor[HTML]{FDFDFE}54.10 \\
 &  &  \multicolumn{1}{r|}{\textit{-Text}\phantom{ }\phantom{ }\phantom{ }}  & \cellcolor[HTML]{FFFCF4}56.77 & \cellcolor[HTML]{F9FBFE}53.26 & \cellcolor[HTML]{5F97F5}9.94 & \cellcolor[HTML]{E8F0FD}48.49 & \cellcolor[HTML]{4788F4}3.27 & \cellcolor[HTML]{FEE7A3}71.36 & \cellcolor[HTML]{BED5FB}36.60 & \cellcolor[HTML]{BDD4FB}36.22 & \cellcolor[HTML]{C8DCFB}39.49 \\
 & \multirow{-3}{*}{\textbf{7B}} &  \multicolumn{1}{r|}{\textit{-Image}} & \cellcolor[HTML]{8FB7F8}23.48 & \cellcolor[HTML]{A8C6F9}30.33 & \cellcolor[HTML]{99BDF9}26.20 & \cellcolor[HTML]{4285F4}1.77 & \cellcolor[HTML]{528FF4}6.39 & \cellcolor[HTML]{DFEAFD}45.95 & \cellcolor[HTML]{FFFCF3}56.85 & \cellcolor[HTML]{C8DBFB}39.29 & \cellcolor[HTML]{A2C3F9}28.78 \\
 \cline{2-12}
 &  & Meme & \cellcolor[HTML]{FFF4D4}62.48 & \cellcolor[HTML]{FFF8E2}60.06 & \cellcolor[HTML]{FEE9AC}\textbf{69.81} & \cellcolor[HTML]{FFFCF3}\textbf{56.99} & \cellcolor[HTML]{FFF1CA}\textbf{64.34} & \cellcolor[HTML]{FEEBB2}68.67 & \cellcolor[HTML]{FFF4D4}62.58 & \cellcolor[HTML]{FEEAAD}69.48 & \cellcolor[HTML]{FFF1CA}\underline{64.30} \\  
 &  &  \multicolumn{1}{r|}{\textit{-Text}\phantom{ }\phantom{ }\phantom{ }}  & \cellcolor[HTML]{FFF1C7}64.78 & \cellcolor[HTML]{FFF2CE}\textbf{63.66} & \cellcolor[HTML]{6CA0F6}13.56 & \cellcolor[HTML]{FAFCFE}53.51 & \cellcolor[HTML]{ADCAFA}31.80 & \cellcolor[HTML]{FEEBB2}68.68 & \cellcolor[HTML]{9CBFF9}27.00 & \cellcolor[HTML]{C2D7FB}37.65 & \cellcolor[HTML]{DCE8FD}45.08 \\
 & \multirow{-3}{*}{\textbf{32B}} &  \multicolumn{1}{r|}{\textit{-Image}} & \cellcolor[HTML]{FFFDF4}56.69 & \cellcolor[HTML]{C4D9FB}38.21 & \cellcolor[HTML]{FAFCFE}53.36 & \cellcolor[HTML]{7BA9F7}17.78 & \cellcolor[HTML]{B0CCFA}32.63 & \cellcolor[HTML]{FFEFC3}65.56 & \cellcolor[HTML]{FFFCF3}56.85 & \cellcolor[HTML]{FFFBEE}57.85 & \cellcolor[HTML]{E4EEFD}47.37 \\
 \cline{2-12}
 &  & Meme & \cellcolor[HTML]{FFF0C5}65.31 & \cellcolor[HTML]{FFF5D9}61.61 & \cellcolor[HTML]{FFF0C4}\underline{65.35} & \cellcolor[HTML]{FCFDFE}54.09 & \cellcolor[HTML]{FFF5D8}61.76 & \cellcolor[HTML]{FEDF85}\underline{76.76} & \cellcolor[HTML]{FFF3D2}62.81 & \cellcolor[HTML]{FEE8A7}\underline{70.57} & \cellcolor[HTML]{FFF1C7}\textbf{64.78} \\
 &  &  \multicolumn{1}{r|}{\textit{-Text}\phantom{ }\phantom{ }\phantom{ }}  & \cellcolor[HTML]{FEEEC0}66.18 & \cellcolor[HTML]{FFF6DD}60.93 & \cellcolor[HTML]{8DB5F8}22.95 & \cellcolor[HTML]{FFFFFC}\underline{55.24} & \cellcolor[HTML]{D0E0FC}41.63 & \cellcolor[HTML]{FEE28F}74.90 & \cellcolor[HTML]{D4E3FC}42.90 & \cellcolor[HTML]{F0F5FE}50.53 & \cellcolor[HTML]{F5F8FE}51.91 \\
\multirow{-9}{*}{\textbf{Qwen}} & \multirow{-3}{*}{\textbf{72B}} &  \multicolumn{1}{r|}{\textit{-Image}} & \cellcolor[HTML]{FBFDFE}53.80 & \cellcolor[HTML]{C3D8FB}37.96 & \cellcolor[HTML]{FFF5D8}61.81 & \cellcolor[HTML]{78A7F7}16.92 & \cellcolor[HTML]{A8C7F9}30.53 & \cellcolor[HTML]{FEEBB3}68.46 & \cellcolor[HTML]{FFF7E0}60.30 & \cellcolor[HTML]{FFFEFA}55.62 & \cellcolor[HTML]{E7F0FD}48.18 \\
\bottomrule
\end{tabular}
}
\caption{F1 scores (in \%) of eight models on the \textit{binary} classification of figurative meaning types in memes from three datasets. 
\textit{-Text} and \textit{-Image} denote the ablation of the text or image modality, respectively.
The \textbf{best} and \underline{second-best} results are shown in bold and underlined, respectively.
Following \citet{liu2022FigMemes}, in \figmemes{} we consider memes with 0 labels as negative cases, and take the best-performing benchmark score as baseline.
Higher scores are highlighted in \colorbox{yellow!70!orange}{orange}, lower scores in \colorbox{blue!50!cyan!40!}{blue}.
Results are averaged over five runs.
}
\label{tab:main_result}
\end{table*}

\begin{table}[ht]
\centering
\resizebox{0.9\linewidth}{!}{
\begin{tabular}{ll|cccc}
\toprule
\textbf{Model} 
& \textbf{Input} & \textbf{Acc.} & \textbf{Prec.} & \textbf{Rec.} & \textbf{F1}  \\
\midrule
\textbf{\textit{random}}  & -- & \cellcolor[HTML]{FFF3CF}50.04 & \cellcolor[HTML]{4385F4}15.57 & \cellcolor[HTML]{FFF3D0}49.88 & \cellcolor[HTML]{87B1F8}23.73 \\\midrule
 &  Meme & \cellcolor[HTML]{FEEAB1}57.60 & \cellcolor[HTML]{8EB6F8}24.63 & \cellcolor[HTML]{FDCE47}83.54 & \cellcolor[HTML]{FEFEFE}38.05 \\
 &   \multicolumn{1}{r|}{\textit{-Text}\phantom{ }\phantom{ }\phantom{ }}  & \cellcolor[HTML]{FEE08B}66.80 & \cellcolor[HTML]{94BAF8}25.30 & \cellcolor[HTML]{FEEAAF}57.86 & \cellcolor[HTML]{E6EFFD}35.20 \\
\multirow{-3}{*}{\textbf{\ayal{}}}  
&  \multicolumn{1}{r|}{\textit{-Image}} & \cellcolor[HTML]{FDD96E}74.06 & \cellcolor[HTML]{ADCAFA}28.36 & \cellcolor[HTML]{FFFAEA}43.51 & \cellcolor[HTML]{DFEAFD}34.34 \\
\midrule
 &  Meme & \cellcolor[HTML]{FEECB7}55.97 & \cellcolor[HTML]{8DB5F8}24.44 & \cellcolor[HTML]{FCCA37}\textbf{87.50} & \cellcolor[HTML]{FFFFFF}38.21 \\
 &   \multicolumn{1}{r|}{\textit{-Text}\phantom{ }\phantom{ }\phantom{ }} & \cellcolor[HTML]{FEE69E}62.08 & \cellcolor[HTML]{87B1F8}23.79 & \cellcolor[HTML]{FEE292}65.04 & \cellcolor[HTML]{E3EDFD}34.84 \\
\multirow{-3}{*}{\textbf{\gemmal{}}} 
&  \multicolumn{1}{r|}{\textit{-Image}} & \cellcolor[HTML]{FEDF84}68.56 & \cellcolor[HTML]{98BDF9}25.86 & \cellcolor[HTML]{FEEDBC}54.83 & \cellcolor[HTML]{E6EEFD}35.15 \\
\midrule
 &  Meme & \cellcolor[HTML]{FEE394}64.69 & \cellcolor[HTML]{ACC9FA}28.24 & \cellcolor[HTML]{FDD04D}82.12 & \cellcolor[HTML]{FFFBF0}\textbf{42.03} \\
 &   \multicolumn{1}{r|}{\textit{-Text}\phantom{ }\phantom{ }\phantom{ }}  & \cellcolor[HTML]{FDDB76}71.91 & \cellcolor[HTML]{B1CCFA}28.76 & \cellcolor[HTML]{FEEEBE}54.31 & \cellcolor[HTML]{FAFCFE}37.60 \\
\multirow{-3}{*}{\textbf{\qwenl{}}} 
&  \multicolumn{1}{r|}{\textit{-Image}} & \cellcolor[HTML]{FDD86D}\textbf{74.27} & \cellcolor[HTML]{B3CEFA}\textbf{29.07} & \cellcolor[HTML]{FFF8E3}45.21 & \cellcolor[HTML]{E8F0FD}35.39 \\
\bottomrule
\end{tabular}
}
\caption{Accuracy, precision, recall  and Micro-F1 (in \%) of models on \textit{multi-label} classification of figurative meanings in \figmemes.
The highest score for each metric is highlighted in \textbf{bold}.
Higher scores are highlighted in \colorbox{yellow!70!orange}{orange}, lower scores in \colorbox{blue!50!cyan!40!}{blue}.}
\label{tab:multi_label}
\end{table}

\section{Results and Discussion}
\label{sec:results}

We begin by focusing on the first prompt setup to discuss about the results of model performance on detecting figurative meaning in Section~\ref{subsec:automatic_evaluation} and on explaining it in Section~\ref{subsec:human_evaluation}. 
The influence of prompt settings will be discussed in Section~\ref{subsec:prompt_sensitivity}.

\subsection{Automatic Evaluation of Detection}
\label{subsec:automatic_evaluation}

\paragraph{Q1: How does model performance vary in detecting different types of figurative meaning?}

Here, we analyze model performance under the original meme input setup.
We start with the \textit{binary} classification tasks (Table~\ref{tab:main_result}), and subsequently, move to \textit{multi-label} classification (Table~\ref{tab:multi_label}), which only applies to \figmemes{}.

As reported in Table~\ref{tab:main_result}, the eight MLLMs show considerable variation: 
Overall, according to their \textit{average} score on eight types, \textbf{the largest model from each family consistently achieves the best results}, yet the differences in overall performance remain marginal. \qwenl{} performs the best among the eight models.

However, on individual label types, the performance gap between models can be more substantial.\footnote{To address potential concerns regarding the variance in model ranking, we conducted a series of McNemar’s tests. Comparing the top two models for each task, we observed that the best-performing model significantly outperformed the second best one ($p < 0.001$) across every experimental seed.}
For instance, on \texttt{metaphor}, the gap between \qwens{} and \qwenm{} is as large as 55.34\%.
Models' performance on \memotion{} and \metmeme{} is more consistent across model sizes, with only minor variations among models from all three series.

\begin{figure*}[ht]
    \centering
    \includegraphics[width=0.98\linewidth]{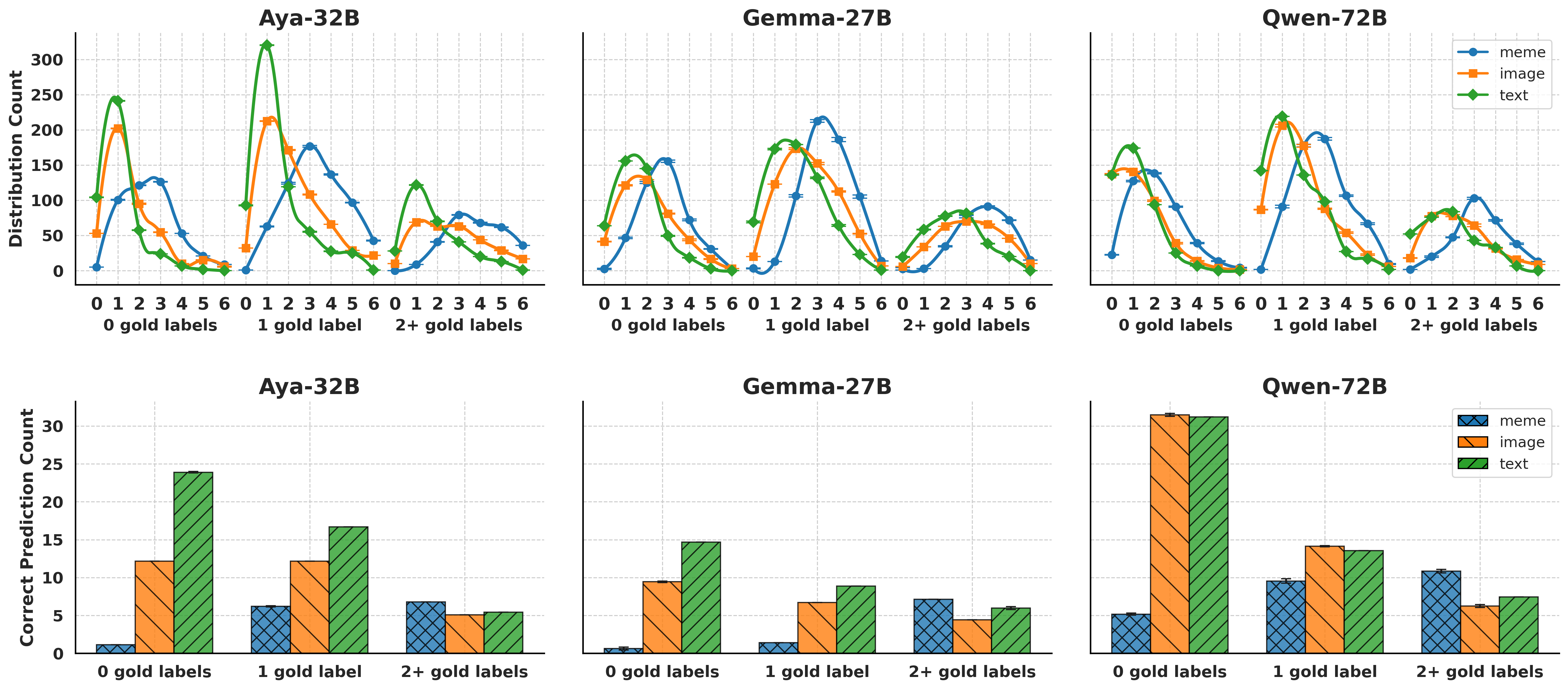}
    \caption{
    Performance of \ayal, \gemmal, and \qwenl{} on the multi-label classification task (only in \figmemes{}) across modalities and meme groups. Error bars indicate standard deviation over 5 runs. 
    \textbf{Top row:} Count of memes with 0, 1, or 2+ figurative types (0, 1, 2+ gold labels) that are predicted to contain from 0 to 6 figurative types.
    \textbf{Bottom row:} Count of memes containing 0, 1, or 2+ figurative types that are fully correctly predicted, where all six predicted labels match the gold labels.
    }
    \label{fig:combined_distribution_and_correct_count}
\end{figure*}

On \figmemes, models generally perform best on \texttt{contrast}, with \ayas{} achieving the highest score of 77.95\%.
In comparison, performance on \texttt{anthropomorphism} is much lower, with the best model, \qwenm, reaching only 56.99\%. 
These trends are consistent with the benchmarks reported by~\citet{liu2022FigMemes}, in which discriminative models also score highest on \texttt{contrast} and lowest on \texttt{anthropomorphism}.

For multi-label classification, we only report the results of the largest model from each family due to space constraints.
Table~\ref{tab:multi_label} shows that all three models surpass the baseline performance, which corresponds to random guessing of each label as positive or negative with a 50/50 probability. 
However, performance differs notably across the four metrics. 
This is largely due to the fact that, on average, each meme has only one positive label while the other five are negative.
For instance, with input of original memes, \gemmal{} achieves the highest recall but the lowest precision among the three models, as the model tend to predict the presence of figurative meaning.
This tendency is generally observed across all three models.

\paragraph{Q2: How much does text, image, or both contribute to the figurativeness of memes?} \label{para:q2}

We begin by examining the \textit{binary classification} task.
On \figmemes{}, we treat the best scores from~\citet{liu2022FigMemes} across modalities and models as a baseline.
As shown in Table~\ref{tab:main_result}, based on the average score across types, performance generally drops when either modality is removed, with the exception of \gemmas{} and \gemmam{}, which score higher on \texttt{exaggeration}, \texttt{anthropomorphism} and \texttt{contrast} in \figmemes{} when one modality is removed.

Interestingly, for all evaluated models, the negative impact of removing the image (\textit{-image}) is more pronounced than removing the text (\textit{-text}) for \texttt{irony} in \figmemes, \texttt{sarcasm} in \memotion, and \texttt{metaphor} in \metmeme{};
whereas for \texttt{allusion}, \texttt{exaggeration}, and \texttt{anthropomorphism} in \figmemes{} the opposite pattern occurs.
This indicates that \textbf{some figurative types are visually grounded while others are primarily conveyed through text}, and MLLMs rely on different modalities to interpret different types of figurative meaning.
Moreover, for \texttt{metaphor} in \figmemes{} and \metmeme, all models (except for \gemmas) exhibit substantial performance drops under both \textit{-Image} and \textit{-Text} conditions, highlighting that detecting metaphors depends not on a single modality alone but on the interplay between text and image.

For the \textit{multi-label classification} task (Table~\ref{tab:multi_label}), micro-recall across all models consistently follows the trend: original meme > \textit{-text} > \textit{-image}, indicating that \textbf{models tend to predict the presence of figurative meaning more often under the full meme input setup}.

To analyze whether models predict more labels due to richer figurative content in a meme (without any ablation) or merely because of a bias toward the meme format, we categorize memes from \figmemes{} by the number of gold labels and count the predicted labels under different modality inputs.
As shown at the top row of Figure~\ref{fig:combined_distribution_and_correct_count}, independent of the number of gold figurative labels a meme carries, the models tend to assign more figurative labels when they see the original meme compared to the ablated versions. 
Notably, among 495 memes without any figurative labels, all three models assign 0 labels to very few cases, while most cases receive 1-3 labels.
Due to this bias, all three models fail to correctly classify memes that do not contain any figurative labels, as reported in the bottom row of Figure~\ref{fig:combined_distribution_and_correct_count}. For such cases, models even perform better with text-only or image-only input, especially for \qwenl.

\begin{figure*}[ht]
    \centering
    \includegraphics[width=\linewidth]{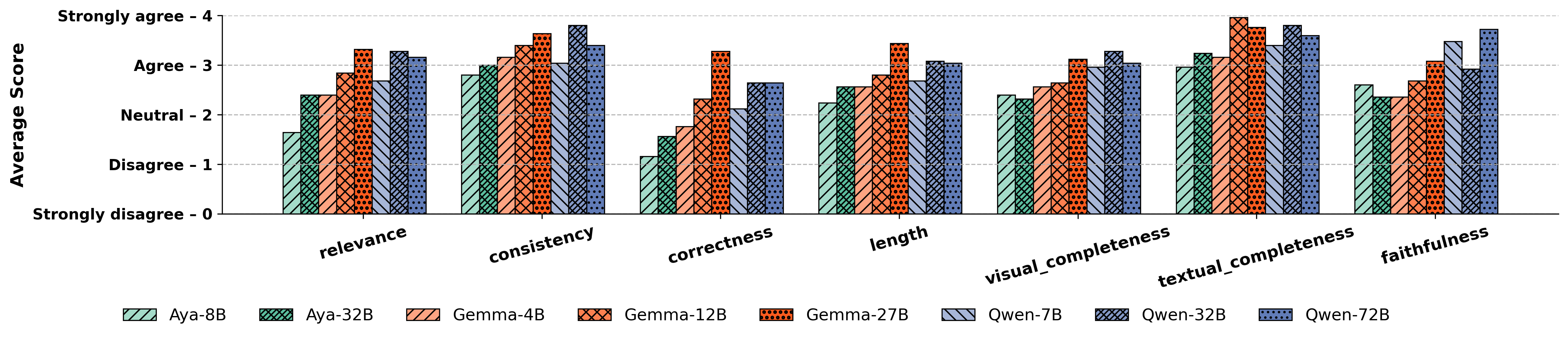}
    \caption{Human evaluation results on model-generated explanations.}
    \label{fig:human_eval_bar_plot}
\end{figure*}

\begin{figure*}[ht]
    \centering
    \begin{subfigure}[b]{0.24\linewidth}
        \centering
        \includegraphics[height=3.3cm]{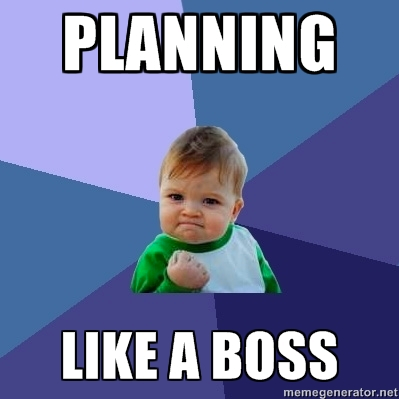}
        \caption{}
        \label{fig:memotion_example_1}
    \end{subfigure}
    \hfill
    \begin{subfigure}[b]{0.24\linewidth}
         \centering
        \includegraphics[height=3.3cm]{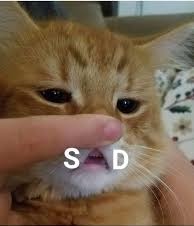}
        \caption{}
        \label{fig:metmeme_exaple_1}
    \end{subfigure}
        \centering
    \begin{subfigure}[b]{0.24\linewidth}
        \centering
        \includegraphics[height=3.3cm]{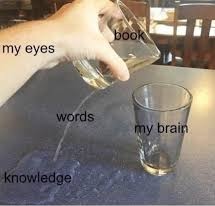}
        \caption{}
        \label{fig:metmeme_exaple_2}
    \end{subfigure}
    \hfill
    \begin{subfigure}[b]{0.24\linewidth}
        \centering
        \includegraphics[height=3.3cm]{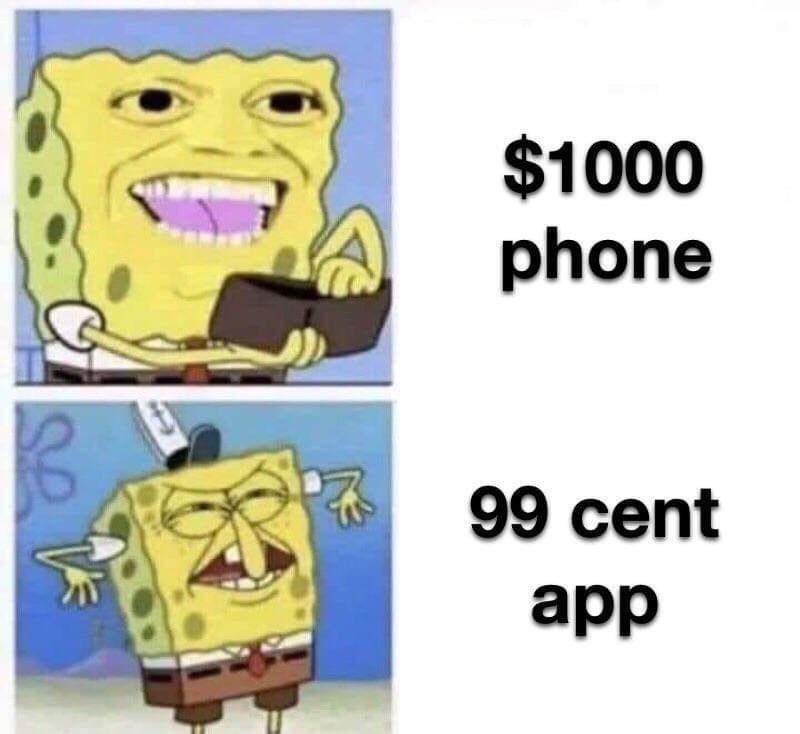}
        \caption{}
        \label{fig:memotion_example_2}
    \end{subfigure}
    \caption{Examples of memes for which MLLMs struggle to generate high-quality explanations.}
    \label{fig:memotion_example}
\end{figure*}

In contrast, for memes containing more than two types of figurative meaning, using the full meme input yields better performance.
This tendency is also observed in other smaller models.
We present the performance of the eight models in Appendix~\ref{subsec:full_analysis}.

\subsection{Human Evaluation of Explanation}
\label{subsec:human_evaluation}

\paragraph{Q3: How effective are the model-provided explanations for its predicted labels and memes?}

We manually evaluate the explanations generated by the eight models following criteria introduced in Section~\ref{subsec:evaluation_setup}.
As shown in Figure~\ref{fig:human_eval_bar_plot}, our comparison reveals an interesting discrepancy: while \qwenl{} was previously found best on automatic metrics, it falls short in human assessment. 
While \gemma{} and \qwen{} models are judged with higher ratings than \aya{} models of comparable size, \qwenl{} surpasses \qwenm{} only on \texttt{faithfulness}, indicating fewer hallucinations, while on other criteria, its performance is roughly equal to or slightly worse than \qwenm{}. Furthermore, for \aya{} and \gemma{}, larger models outperform smaller ones.

Focusing on the internal coherence between predicted labels and explanations, we find that \textbf{all models perform better on \texttt{consistency} than \texttt{relevance}}. 
Model-generated explanations often lack substance: despite aligning with the predicted labels, they frequently fail to provide concrete evidence for the specific figurative meaning, such as identifying the target of sarcasm or the underlying metaphor.

Human annotators rated the models highest on \texttt{textual completeness}, indicating a strong consensus that the generated explanations successfully capture the key textual elements essential to the memes' figurative meanings.

By comparison, the models perform worse in capturing visual components. 
Most models perform worst on \texttt{correctness}: \ayas{}, 
\ayal{}, and \gemmas{} all score below 2, suggesting that human annotators generally disagree with the models' interpretations of the memes intent.

\paragraph{Q4: What are the common types of errors in model explanations?} 

As we found in Section~\ref{para:q2}, MLLMs' strong bias toward predicting the presence of figurative meaning in meme formats results in two key limitations in their explanations.
One is \textbf{label-explanation inconsistency}: for example, Figure~\ref{fig:memotion_example_1} is a motivating meme without sarcastic meaning. \qwenl{} labels it as \textit{sarcastic}, but in its explanation, it contradicts the label and asserts that there is no sarcasm. The other is \textbf{over-interpretation}, where the model attempts to identify an uncommon or exaggerated aspect to justify a positive label.
For instance, \gemmam{} labels Figure~\ref{fig:memotion_example_1} as \textit{sarcastic} and explains that it ``utilizes the \texttt{Success Kid} meme paired with the phrase \texttt{Planning Like A Boss}, creating a strong sense of sarcasm because the baby's determined expression contrasts humorously with the often chaotic reality of planning.''

Mismatches between model-generated explanations and the memes are largely due to failure in visual information extraction rather than in textual processing.
These issues typically arise from either an \textbf{incorrect depiction} of the image or a \textbf{neglect of visual cue}.
For the former, \ayal{} misinterprets Figure~\ref{fig:metmeme_exaple_1} as ``[...] The finger over the cat's mouth is a visual metaphor for silence, implying that the cat should keep quiet or be hushed, thereby attributing human-like behavior to the animal.''
This is inaccurate, as people usually use a vertical finger in front of the mouth to signal silence, whereas the image shows a finger lifting the cat’s nose, causing its mouth to open and make an ``A''.
For the latter, although all models predict the label \textit{metaphoric} correctly, but fail to interpret Figure~\ref{fig:metmeme_exaple_2}, because they overlook the crucial visual cue, the water spilling away before it reaches the glass, and thus misread the meme as ``[...] symbolizing how information is absorbed and processed'', rather than the intended meaning of failing to do so.

We also find that models struggle with memes depicting everyday human behavior. Figure~\ref{fig:memotion_example_2} satirizes people who are willing to spend \$1,000 on a phone but feel annoyed when an app is not free, even if it costs only 99 cents.
This demonstrates the common psychological pattern where consumers rationalize big purchases but resist minimal charges, yet half of the models fail to interpret it.
This type of error can be described as a \textbf{behavioral reasoning error}, where the model misses the social or psychological irony underlying the meme due to lack of social experience, which is also pointed out by ~\citet{chen2024survey} and~\citet{mathur2024Advancing}.

\begin{figure}[t]
    \centering
    \begin{subfigure}[b]{0.48\linewidth}
        \centering
        \includegraphics[height=3.3 cm]{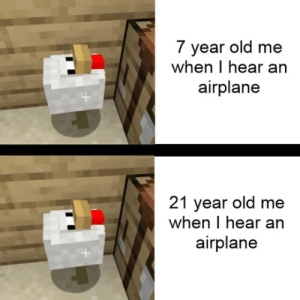}
        \caption{}
        \label{fig:ambiguous1}
    \end{subfigure}
    \hfill
    \begin{subfigure}[b]{0.48\linewidth}
        \centering
        \includegraphics[height=3.3 cm]{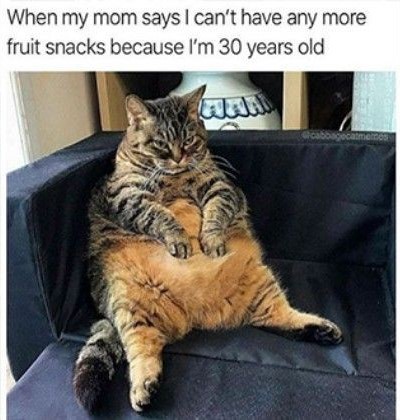}
        \caption{}
        \label{fig:ambiguous2}
    \end{subfigure}
    \caption{Two memes from \memotion{} with gold label ``not sarcastic''.}
    \label{fig:ambiguous}
\end{figure}

\paragraph{Q5: Can explanations be valid for ``wrong'' labels?}

When qualitatively evaluating model-generated labels and explanations, we observed that models can assign labels that are different from the gold label while still providing reasonable and evidence-based explanations. For example, both memes in Figure~\ref{fig:ambiguous} are annotated with \textit{not sarcastic}, yet \gemmal{} classifies the first as \textit{sarcastic}, describing it as ``creating a sarcastic tone about the unchanging nature of this specific behavior and subtly mocking the continued enthusiasm.'' and \qwenl{} labels the second as \textit{sarcastic}, explaining it as ``mocking the concept of age-related restrictions on simple pleasures''. Both annotators (strongly) agree that these explanations fit well with the seven human evaluation criteria. These cases highlight the inherent subjectivity in interpretation of figurative meaning and indicate that minority-vote labels are not necessarily errors.

\begin{figure}[t]
    \centering
    \includegraphics[width=\linewidth]{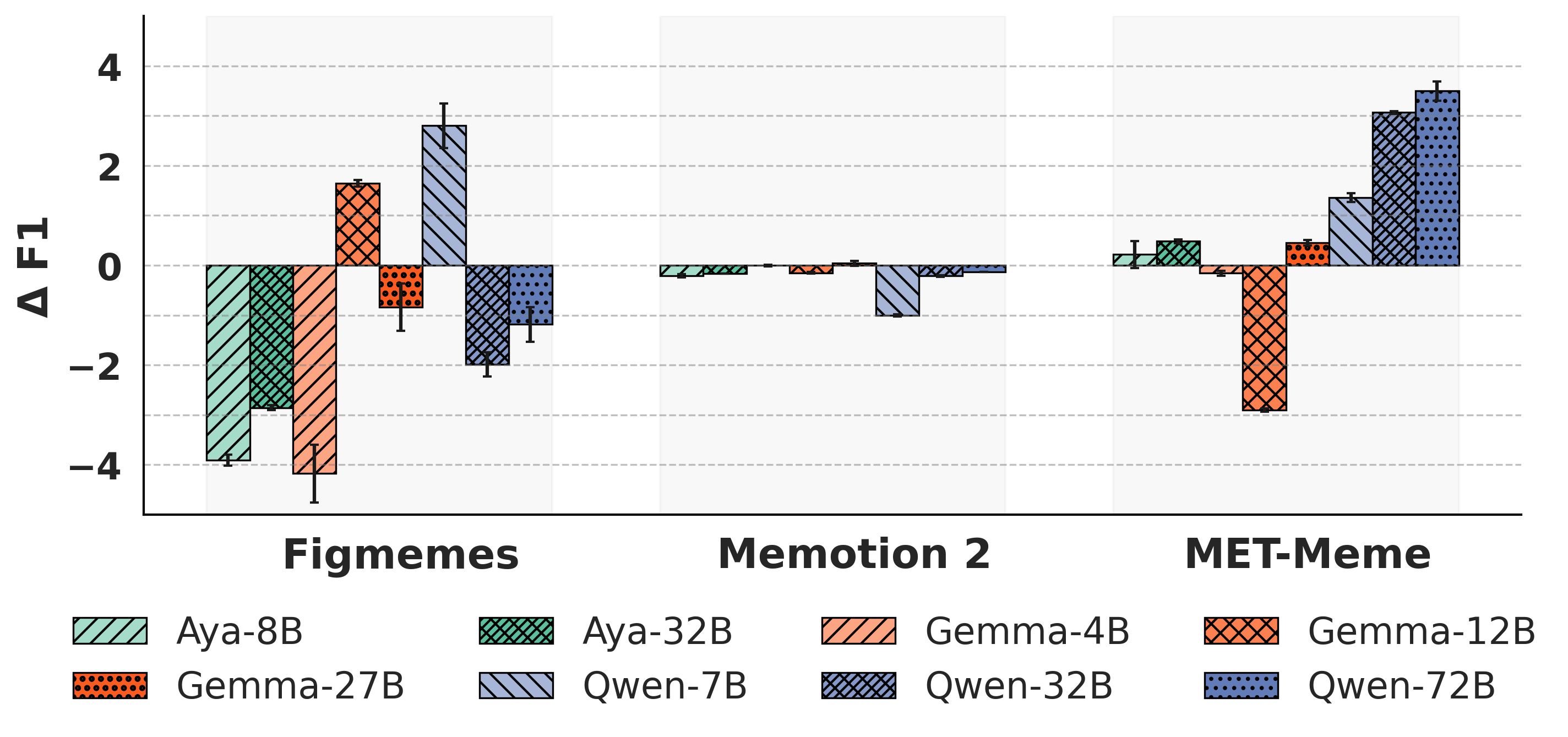}
    \caption{Difference in macro-F1 (in \%) between prompting score and prompting label setting for eight models, averaged over five seeds.
    Bars with positive values indicate that prompting score improved performance; bars with negative values indicate a decrease.
    Results cover binary classification tasks across three datasets.} 
    \label{fig:prompt_sensitivity}
\end{figure}

\subsection{Prompt Sensitivity}
\label{subsec:prompt_sensitivity}

\paragraph{Q6: How robust are models to different prompts?}

Since MLLMs are highly sensitive to prompt design, we validate our results with two prompting strategies (see Section~\ref{subsec:task_setup}).
The full prompt template and the corresponding experimental results are provided in  Appendix~\ref{subsec:second_prompt}.

Although F1 scores under the same prompt setting show little variation (less than 1\% standard deviation), model performance under different prompt settings varies moderately. Figure~\ref{fig:prompt_sensitivity} quantifies the gap between results of the binary classification task under two prompting setups. Overall, the gap of performance of all models is within $\pm4\%$. Models demonstrate higher robustness on \memotion{} than on \figmemes{} and \metmeme{}.
\gemmal{} is the most consistent model, with performance variations across the three datasets less than 1\%.
There is no clear evidence that any particular prompt setting consistently benefits a specific model in all setups; performance still largely depends on the dataset.

\section{Conclusion} \label{sec:conclusion}
We explore the ability of current generative multimodal models to \textit{detect and explain} figurative meaning in internet memes.
We evaluate in total eight models from three families, \aya, \gemma, and \qwen{} with different sizes, and analyze the influence of input modalities (text, image, both) quantitatively and qualitatively. 

Larger models generally achieve better results on the binary classification task. Textual components contribute more to detecting \texttt{irony} and \texttt{metaphor}, whereas visual components are more critical for \texttt{allusion}, \texttt{exaggeration}, and \texttt{anthropomorphism}. All evaluated models exhibit a bias toward predicting the presence of figurative meaning in meme format input, which is typically composed of images with embedded text, suggesting a higher reliance on surface characteristics.

Importantly, our human evaluation reveals several deficiencies of model-generated \textit{explanations}. Across all models, identifying the core visual components that convey figurative meaning is harder than identifying the textual ones.

When model-generated explanations fail to capture the figurative meaning of a meme, it is largely due to failures in visual information extraction rather than in textual processing.
Furthermore, models sometimes assign labels that differ from the gold annotation while still generating valid, evidence-based explanations, suggesting that minority-vote labels are not necessarily errors but may reflect the inherent subjectivity of figurative meaning interpretation.

Ultimately, our findings highlight the need for future research to move beyond simple label prediction toward developing more sophisticated multimodal architectures that can ensure reasoning faithfulness and a deeper sensitivity to the complex socio-psychological contexts inherent in meme communication.

\section*{Limitations}

\paragraph{Data and Benchmark Limitations}
The three-year release history of the evaluation datasets poses a risk of data contamination, potentially inflating performance if models encountered these samples during pre-training.
Furthermore, the limited variety of meme formats, imbalanced label distribution (e.g., sparse figurative types in \figmemes{}), and inconsistent annotation schemes across datasets collectively constrain the representativeness of our findings.
These factors complicate fair comparisons and may bias the assessment of models' generalization capabilities across diverse communicative contexts.

\paragraph{Robustness to Prompt Specification}
Despite evaluating two prompt templates across five experimental runs, potential sensitivity to prompt specification remains. For example, it is unclear how consistency might be affected by variations in output format (e.g., XML vs. JSON) or subtle shifts in instructional wording.
Furthermore, while our continuous scoring-type prompt (0--1) provides fine-grained scalar feedback, a discrete Likert-like scale might introduce different inductive biases.

\newpage

\section*{Ethical Considerations}

\paragraph{Hateful content in \figmemes}
To assess potential ethical concerns, we randomly sampled 30 memes from \figmemes{}, finding that 17 could be considered potentially offensive, targeting groups such as religious or social communities. While the dataset contains sensitive content, all annotations in our study were conducted with care, and no annotator reported any psychological distress or harm during the annotation process. We emphasize that our study focuses on understanding figurative language in these memes for research purposes and does not promote or disseminate harmful content. Access to the dataset is controlled, and all participants are expected to follow ethical research practices when using it.

\paragraph{Use of AI Assistants.}
The authors acknowledge the use of ChatGPT solely for correcting grammatical errors, enhancing the coherence of the final manuscripts, and providing assistance with coding. 

\section*{Acknowledgements}  \label{sec:acknowledgements}

We thank MaiNLP lab members Felicia Körner, Philipp Mondorf and Andreas Säuberli for giving feedback on earlier drafts of this paper, as well as to the anonymous reviewers for their feedback.
This work is supported by the KLIMA-MEMES project funded by the Bavarian Research Institute for Digital Transformation (bidt), an institute of the Bavarian Academy of Sciences and Humanities.
The authors are responsible for the content of this publication.

\section*{Bibliographical References}\label{sec:reference}

\bibliographystyle{lrec2026-natbib}
\bibliography{mypaper, anthology}

\bibliographystylelanguageresource{lrec2026}

\newpage
\appendix

\section*{Appendices}
\section{Image Preprocessing} \label{sec:image_preprocessing}

\begin{figure}[ht]
    \centering
    \def\imgwidth{0.42\linewidth}
    
    \begin{tabular}{cc}
        \textbf{Original} & \textbf{Inpainted} \\
        \includegraphics[width=\imgwidth]{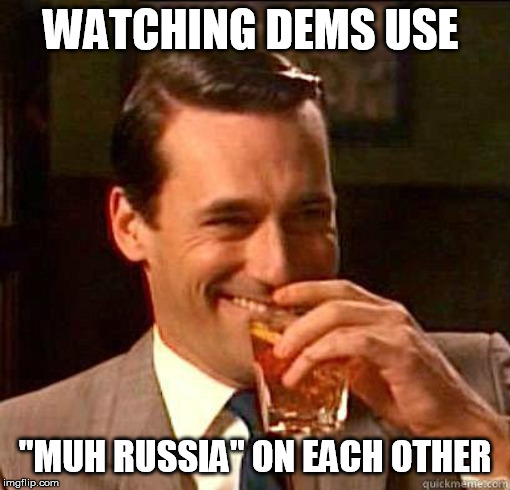} & 
        \includegraphics[width=\imgwidth]{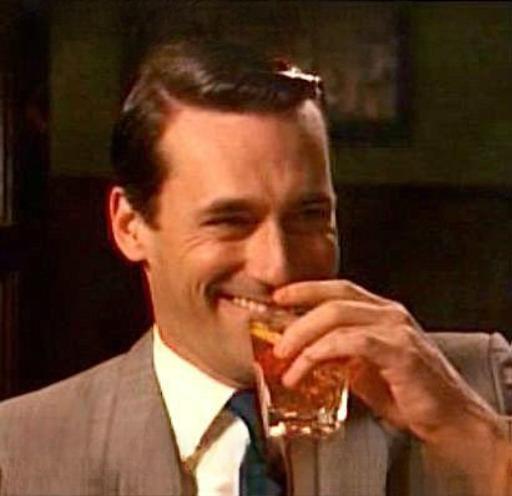} \\
        \specialrule{0em}{1pt}{1pt}
        
        \includegraphics[width=\imgwidth]{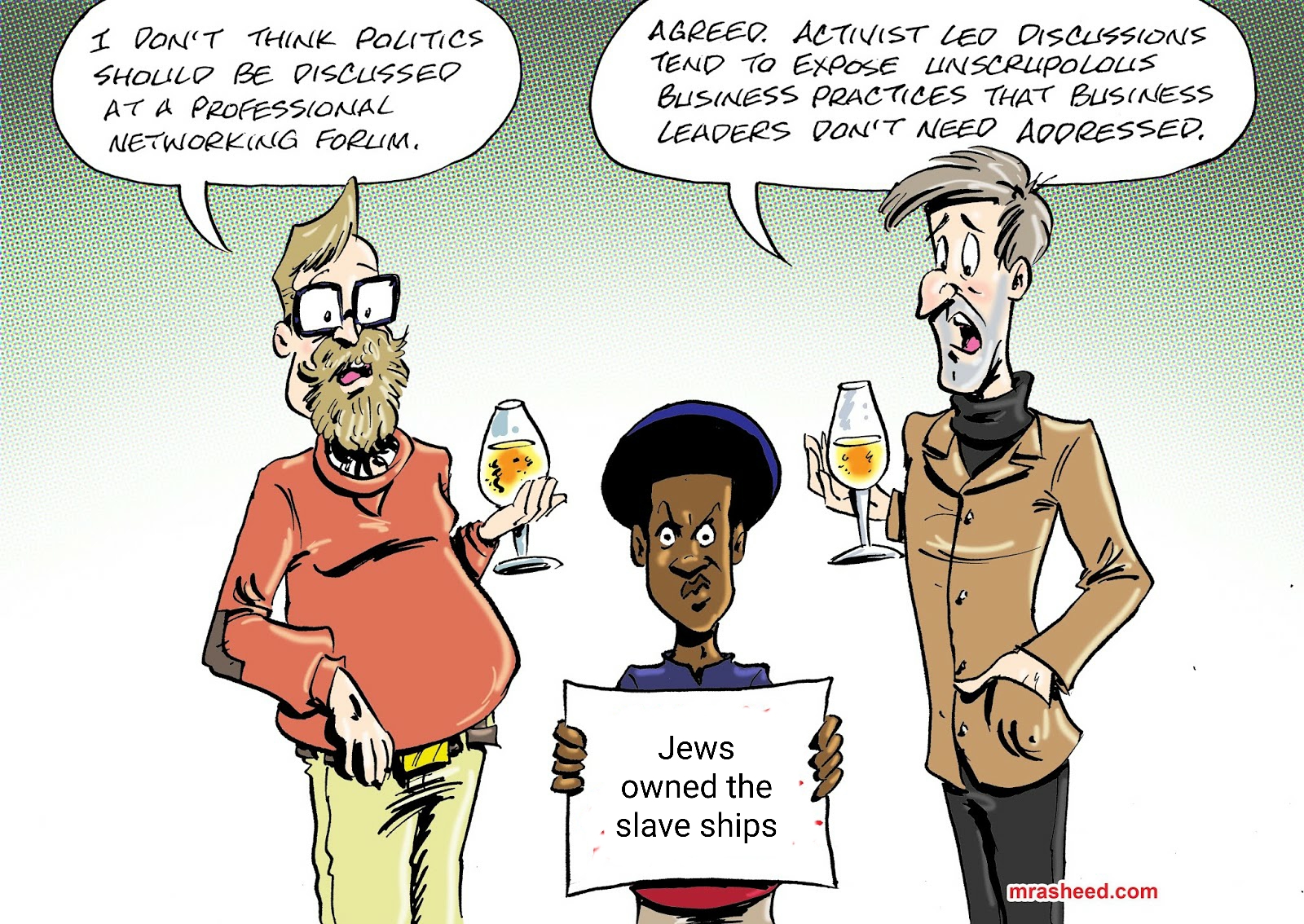} & 
        \includegraphics[width=\imgwidth]{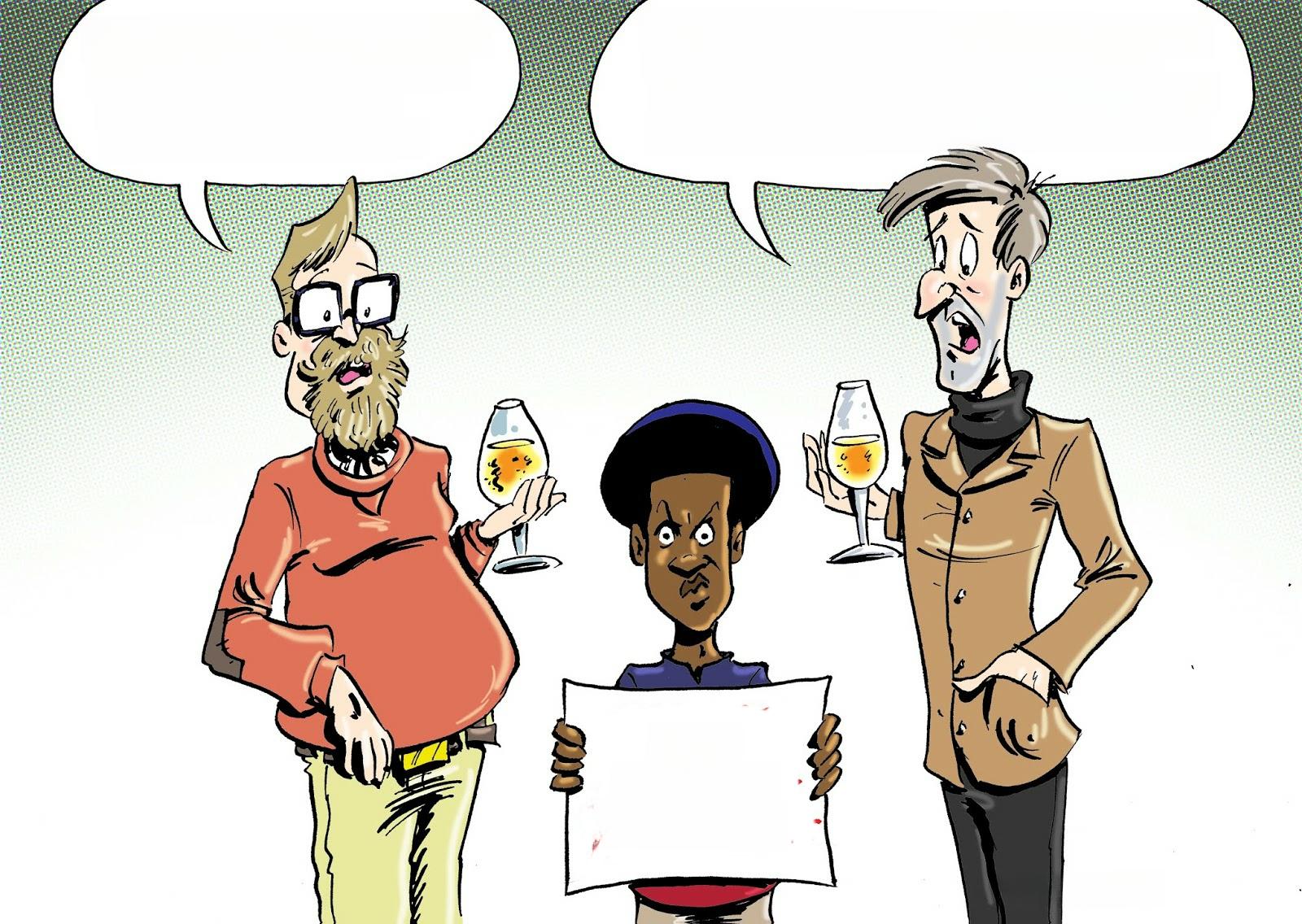} \\
        \specialrule{0em}{1pt}{1pt}
        
        \includegraphics[width=\imgwidth]{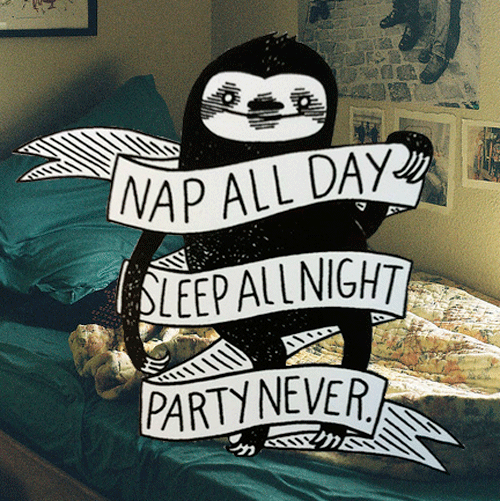} & 
        \includegraphics[width=\imgwidth]{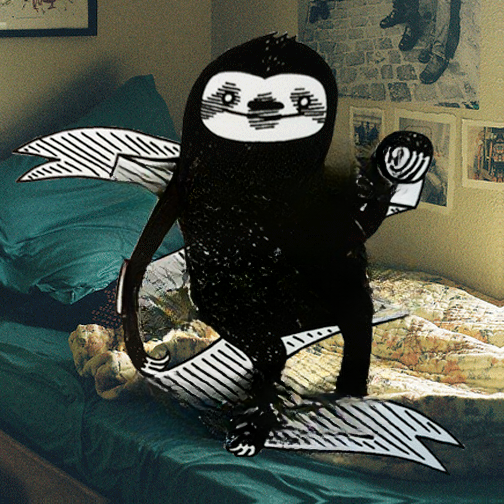} \\
    \end{tabular}
    
    \caption{Examples of automatic preprocessing by inpainting text. The left column shows the original meme images, while the right column displays the results after text removal using PaddleOCR and LaMa.}
    \label{fig:good_examples}
\end{figure}

\begin{figure*}[ht]
    \centering
    \def\imgwidth{0.30\linewidth}
    
    \begin{tabular}{ccc}
        \textbf{Original} & \textbf{Inpainted} & \textbf{Corrected} \\
        \includegraphics[width=\imgwidth]{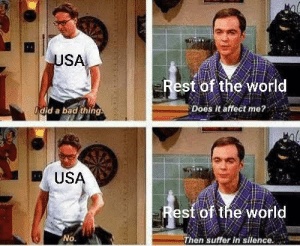} & 
        \includegraphics[width=\imgwidth]{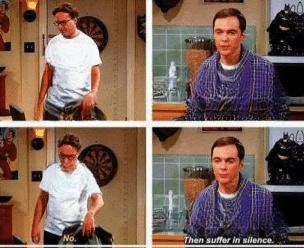} &
        \includegraphics[width=\imgwidth]{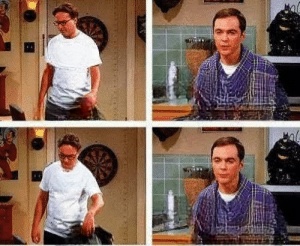}
    \end{tabular}
    
    \caption{Examples of automatic and manual text removal. For cases where low image resolution hindered automated OCR detection, the original memes were manually edited to ensure a clean, text-free input for the model.}
    \label{fig:bad_examples}
\end{figure*}

To ensure the complete removal of textual information from the input—rather than merely relying on prompt-based attention control—we preprocessed the original images.
Specifically, we employed PaddleOCR~\citep{cui2025PaddleOCR} to detect text and mask it, and then use LaMa~\citep{suvorov2021Resolutionrobust} for inpainting.
In most cases, the text was successfully eliminated, as illustrated in Figure~\ref{fig:good_examples}. For a small subset of low-resolution images where automated processing was suboptimal, we manually refined the results to ensure data quality as shown in Figure~\ref{fig:bad_examples}.

\section{Implements Detail and Additional Results} \label{sec:additional_results}

\subsection{Macro-Averaged Results for Multi-label Classification} \label{subsec:macro_score}

For completeness, we also report macro-averaged accuracy, precision, recall and F1 for multi-label classification on \figmemes{} in Table~\ref{tab:multi_label_macro}. Unlike micro-averaging, macro-averaging computes scores independently for each label and takes their unweighted mean, which may be less stable in the presence of label sparsity. These scores are provided as supplementary reference alongside the micro-averaged results reported in the main paper.

\begin{table}[ht]
\centering
\resizebox{0.9\linewidth}{!}{
\begin{tabular}{ll|cccc}
\toprule
\textbf{Model} 
& \textbf{Input} & \textbf{Acc.} & \textbf{Prec.} & \textbf{Rec.} & \textbf{F1}  \\
\midrule
 &  Meme & 57.60 & 24.41 & 82.72 & 37.27 \\
 &   \multicolumn{1}{r|}{\textit{-Text}\phantom{ }\phantom{ }\phantom{ }}  & 66.80 & 25.73 & 61.90 & 34.71 \\
\multirow{-3}{*}{\textbf{\ayal{}}}  
&  \multicolumn{1}{r|}{\textit{-Image}} & 74.06 & 26.90 & 39.98 & 30.23 \\
\midrule

 &  Meme & 87.50	& 37.88	& 55.97	& 24.57
 \\
 &   \multicolumn{1}{r|}{\textit{-Text}\phantom{ }\phantom{ }\phantom{ }} & 69.09 & 34.82	& 62.08 & 24.12 
 \\
\multirow{-3}{*}{\textbf{\gemmal{}}} 
&  \multicolumn{1}{r|}{\textit{-Image}} & 51.24 & 31.44	& 68.56	& 23.70	
 \\
\midrule
 &  Meme & 81.19 & 41.73	& 64.69	& 28.69
 \\
 &   \multicolumn{1}{r|}{\textit{-Text}\phantom{ }\phantom{ }\phantom{ }}  & 60.06 & 35.90	& 71.91	& 28.71	
 \\
\multirow{-3}{*}{\textbf{\qwenl{}}} 
&  \multicolumn{1}{r|}{\textit{-Image}} & 41.90 & 31.92	& 74.27	& 28.57	
 \\
\bottomrule
\end{tabular}
}
\caption{Macro-averaged accuracy, precision, recall and F1 (in \%) of models on \textit{multi-label} classification of figurative meanings in \figmemes.
}
\label{tab:multi_label_macro}
\end{table}

\subsection{Main Results with Standard Deviation} \label{subsec:result_with_std}

To account for potential performance variations, we conducted five independent trials for the experiments. The resulting average scores, along with the standard deviations which indicate the robustness of our approach, are reported in Table ~\ref{tab:main_result_std}.

\subsection{Model Setup}
\label{subsec:model_setup}

We evaluate the following vision-language models: Aya-Vision
\footnote{\url{https://huggingface.co/collections/CohereLabs/cohere-labs-aya-vision-67c4ccd395ca064308ee1484}} 
\citep{ustun-etal-2024-aya}, Gemma 3%
\footnote{\url{https://huggingface.co/collections/google/gemma-3-release-67c6c6f89c4f76621268bb6d}} 
\citep{team2025Gemma}, and Qwen2.5-VL%
\footnote{\url{https://huggingface.co/collections/Qwen/qwen25-vl-6795ffac22b334a837c0f9a5}} 
\citep{bai2025Qwen25VLa}.
For \gemma\ and \qwen, we test their instruction-tuned versions.

All models ran locally on NVIDIA A100 and H200 GPUs with the \texttt{vLLM}\footnote{\url{https://github.com/vllm-project/vllm}} framework ~\citep{10.1145/3600006.3613165} for efficient and consistent inference. Table~\ref{tab:sampling_params} summarizes the decoding hyperparameters used during inference across all models. Each experiment is repeated 5 times using fixed random seeds: 42, 52, 62, 72, and 82. 

\begin{table}[ht]
\centering
\small
\begin{tabular}{ll}
\toprule
\textbf{Parameter} & \textbf{Value} \\
\midrule
Temperature & 0.7 \\
Top-\textit{p} & 0.1 \\
Repetition Penalty & 1.05 \\
Max Tokens & 512 \\
\bottomrule
\end{tabular}
\caption{Sampling parameters used during inference.}
\label{tab:sampling_params}
\end{table}

\subsection{Performance by Label Count} \label{subsec:full_analysis}

\begin{figure*}[ht] %
    \centering
    \includegraphics[width=0.85\textwidth]{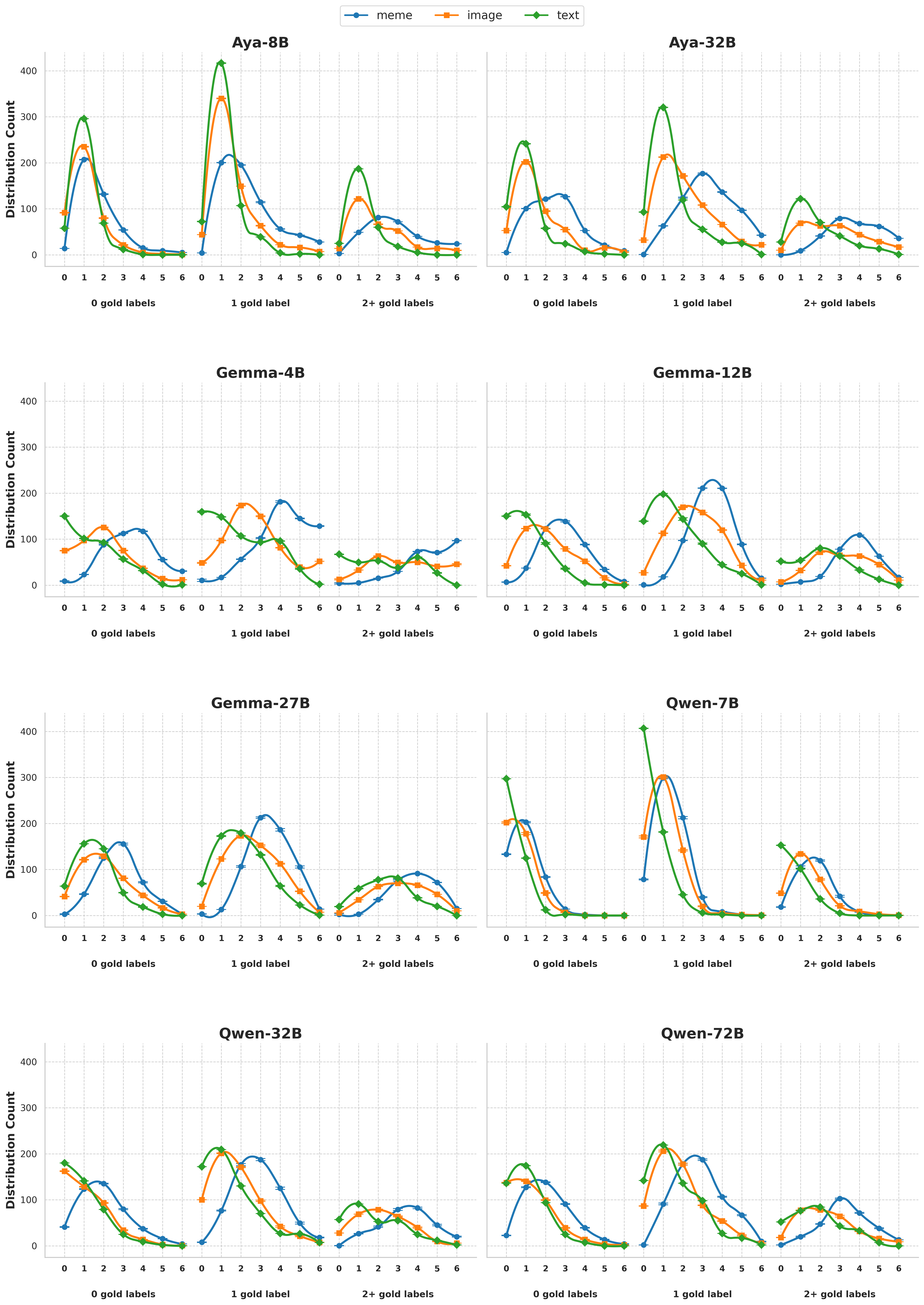}
    \caption{Distribution of predicted label counts across eight models. Each subplot shows the frequency of models predicting 0 to 6 labels, grouped by the actual number of gold labels (0, 1, and 2+). }
    \label{fig:label_distribution_8models}
\end{figure*}

\begin{figure*}[ht]
    \centering
    \includegraphics[width=0.85\textwidth]{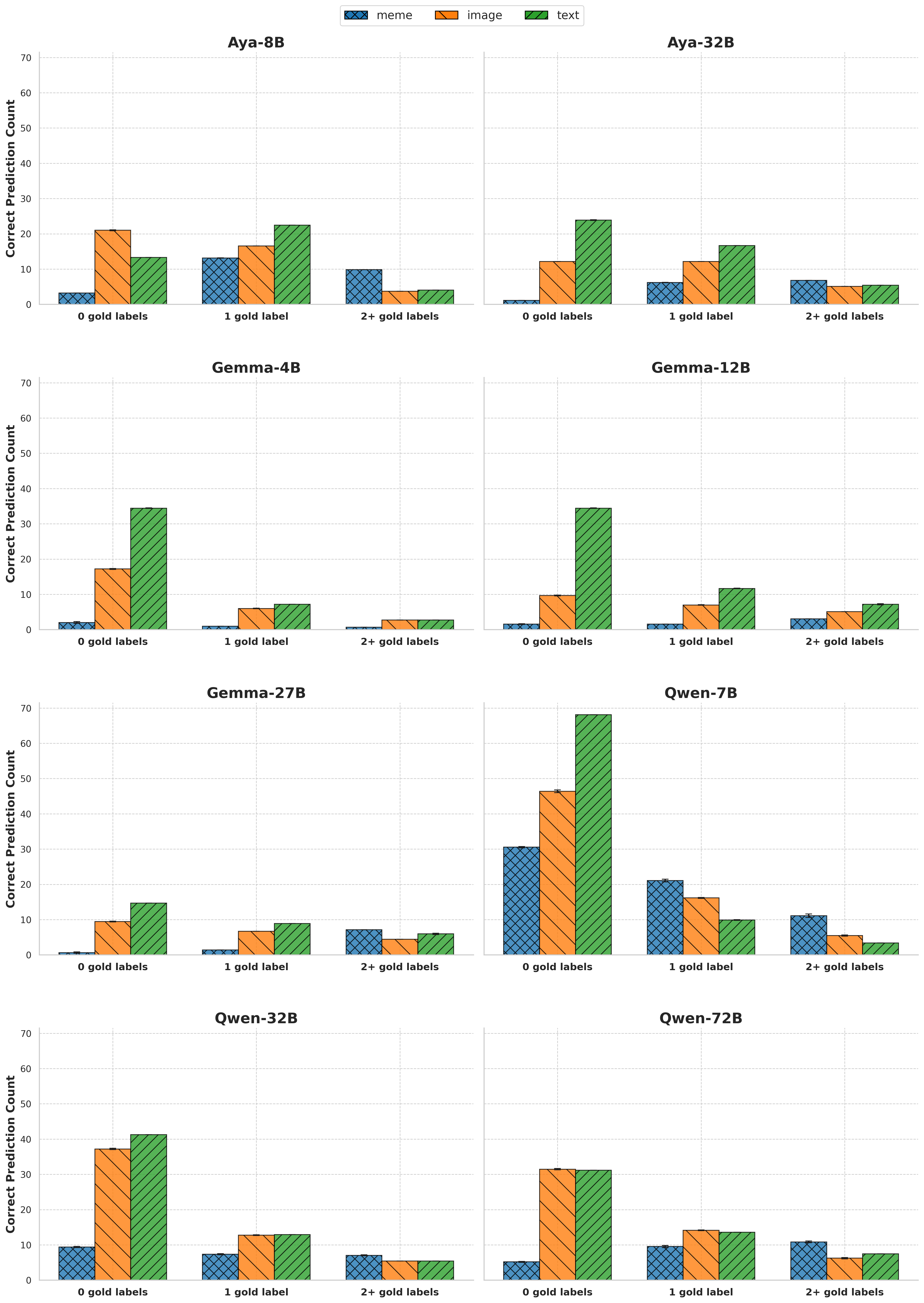}
    \caption{Correct prediction counts for eight models across different label complexities. The bar charts represent the Mean and Standard Deviation (error bars) over five experimental seeds.}
    \label{fig:correct_count_8models}
\end{figure*}

Due to space constraints, Figure~\ref{fig:combined_distribution_and_correct_count} in the main text only presents the performance of the largest model from each series. Here, we provide the complete results for all tested models, as shown in Figure~\ref{fig:label_distribution_8models} and Figure~\ref{fig:correct_count_8models}.

\subsection{Second Prompt Setting Results} \label{subsec:second_prompt}

In the second prompt setting, we prompt the model to provide a continuous, probability-like value between 0 and 1 to indicate the degree to which
the input expresses a specific type of figurative meaning.

\begin{center}
\begin{tikzpicture}
\node[
    draw,
    rectangle,
    rounded corners,
    fill=gray!10,
    inner sep=8pt,
    text width=0.9\linewidth,
    font=\small,
    align=left
] {
\ttfamily
System:\\
You are an expert linguistic annotator.\\[0.5\baselineskip]

User:\\
You are asked to annotate whether the input\\
\ \ \ \ contains a \{FIGURATIVE TYPE\} expression\\
\ \ \ \ or not.\\[0.5\baselineskip]

Definition:\\
\ \ \ \ \{FIGURATIVE TYPE\}: \{DEFINITION\}\\[0.5\baselineskip]

\{INPUT\}\\[0.5\baselineskip]

Please explain why you are assigning this\\
\ \ \ \ label. Your explanation should clearly\\
\ \ \ \ justify your choice and reference the\\
\ \ \ \ relevant visual and/or textual compon-\\
\ \ \ \ ents in the input.\\[0.5\baselineskip]

Return your answer ONLY as a JSON object\\
\ \ \ \ in this exact format:\\[0.5\baselineskip]

\{\\
\ \ "\textbf{score}": \\
\ \ \ \ \ \ \{FIGURATIVE TYPE\}: \\
\ \ \ \ \ \ \ \ \ \ \textbf{probability\_between\_0\_and\_1},\\
\ \ "explanation": \{generated\_explanation\}\\
\}\\[0.5\baselineskip]

ONLY return a valid JSON object in the \\
\ \ \ \ exact format above.
};
\end{tikzpicture}
\end{center}

The averaged score over 5 runs with standard deviations of experiments with second prompt template setup in Table~\ref{tab:main_result_std_second}. 

\begin{table*}[t]
\centering
\resizebox{0.98\linewidth}{!}{
\begin{tabular}{lll|*{6}{>{\raggedleft\arraybackslash}p{1.5cm}}|c|c}
\toprule
 &  &  & \multicolumn{6}{c|}{\textbf{\figmemes}} & \multicolumn{1}{c|}{\textbf{\memotion}} & \multicolumn{1}{c}{\textbf{\metmeme}} \\
\multirow{-2}{*}{\textbf{Model}} & 
\multirow{-2}{*}{\textbf{Size}} & 
\multirow{-2}{*}{\textbf{Input\phantom{ }\phantom{ }\phantom{ }\phantom{ }\phantom{ }\phantom{ }}} & 
\multicolumn{1}{c}{\textbf{Allusion}} & 
\multicolumn{1}{c}{\textbf{Exag.}} & 
\multicolumn{1}{c}{\textbf{Irony}} & 
\multicolumn{1}{c}{\textbf{Anthrop.}} & 
\multicolumn{1}{c}{\textbf{Metaphor}} & 
\multicolumn{1}{c|}{\textbf{Contrast}} & 
\multicolumn{1}{c|}{\textbf{Sarcasm}} & 
\multicolumn{1}{c}{\textbf{Metaphor}} \\
\midrule
\multicolumn{3}{c|}{\textit{\textbf{Baseline}}} & 52.32 & 44.00 & 49.77 & 41.76 & 44.87 & 56.91 & -- & -- \\\midrule
 &  & Meme & 70.04\scriptsize{±0.00} & 58.63\scriptsize{±0.00} & 56.06\scriptsize{±0.06} & 51.94\scriptsize{±0.00} & 50.62\scriptsize{±0.15} & 77.95\scriptsize{±0.00} & 63.48\scriptsize{±0.00} & 66.66\scriptsize{±0.26} \\
 &  &  \multicolumn{1}{r|}{\textit{-Text}\phantom{ }\phantom{ }\phantom{ }} & 65.44\scriptsize{±0.00} & 58.52\scriptsize{±0.12} & 10.53\scriptsize{±0.00} & 48.39\scriptsize{±0.19} & 35.91\scriptsize{±0.22} & 59.99\scriptsize{±0.38} & 49.64\scriptsize{±0.00} & 54.16\scriptsize{±0.22} \\
 & \multirow{-3}{*}{\textbf{8B}} &  \multicolumn{1}{r|}{\textit{-Image}} & 34.78\scriptsize{±0.00} & 30.16\scriptsize{±0.00} & 57.07\scriptsize{±0.00} & 8.47\scriptsize{±0.00} & 40.09\scriptsize{±0.00} & 23.46\scriptsize{±0.00} & 63.01\scriptsize{±0.01} & 57.28\scriptsize{±0.13} \\
 \cline{2-11}
 &  & Meme & 66.87\scriptsize{±0.00} & 61.01\scriptsize{±0.00} & 61.31\scriptsize{±0.03} & 54.67\scriptsize{±0.00} & 54.97\scriptsize{±0.00} & 70.85\scriptsize{±0.07} & 63.37\scriptsize{±0.00} & 66.77\scriptsize{±0.04} \\
 &  &  \multicolumn{1}{r|}{\textit{-Text}\phantom{ }\phantom{ }\phantom{ }} & 67.48\scriptsize{±0.00} & 59.45\scriptsize{±0.00} & 42.27\scriptsize{±0.17} & 51.70\scriptsize{±0.00} & 39.48\scriptsize{±0.22} & 60.13\scriptsize{±0.07} & 47.22\scriptsize{±0.09} & 56.03\scriptsize{±1.31} \\
\multirow{-6}{*}{\textbf{Aya}}  & \multirow{-3}{*}{\textbf{32B}} &  \multicolumn{1}{r|}{\textit{-Image}} & 48.99\scriptsize{±0.00} & 37.66\scriptsize{±0.20} & 59.09\scriptsize{±0.04} & 9.30\scriptsize{±0.00} & 37.85\scriptsize{±0.05} & 63.53\scriptsize{±0.00} & 59.82\scriptsize{±0.03} & 50.11\scriptsize{±1.62} \\
\midrule\midrule
 &  & Meme & 70.17\scriptsize{±0.49} & 53.94\scriptsize{±0.14} & 62.70\scriptsize{±0.06} & 51.37\scriptsize{±0.36} & 53.22\scriptsize{±0.30} & 56.71\scriptsize{±0.11} & 63.47\scriptsize{±0.01} & 66.91\scriptsize{±0.05} \\
 &  &  \multicolumn{1}{r|}{\textit{-Text}\phantom{ }\phantom{ }\phantom{ }}  & 58.13\scriptsize{±0.33} & 58.36\scriptsize{±0.09} & 31.19\scriptsize{±0.51} & 52.06\scriptsize{±0.19} & 54.05\scriptsize{±0.13} & 61.84\scriptsize{±0.06} & 62.05\scriptsize{±0.05} & 60.02\scriptsize{±0.35} \\
 & \multirow{-3}{*}{\textbf{4B}} &  \multicolumn{1}{r|}{\textit{-Image}} & 33.29\scriptsize{±0.06} & 43.46\scriptsize{±0.10} & 58.33\scriptsize{±0.07} & 3.86\scriptsize{±0.73} & 47.19\scriptsize{±0.09} & 61.20\scriptsize{±0.13} & 63.08\scriptsize{±0.00} & 61.42\scriptsize{±0.20} \\
 \cline{2-11}
 &  & Meme & 60.27\scriptsize{±0.05} & 56.09\scriptsize{±0.05} & 63.01\scriptsize{±0.04} & 53.00\scriptsize{±0.00} & 42.87\scriptsize{±0.19} & 57.48\scriptsize{±0.04} & 63.54\scriptsize{±0.02} & 67.81\scriptsize{±0.01} \\
 &  &  \multicolumn{1}{r|}{\textit{-Text}\phantom{ }\phantom{ }\phantom{ }}  & 64.06\scriptsize{±0.11} & 56.96\scriptsize{±0.04} & 45.54\scriptsize{±0.00} & 53.94\scriptsize{±0.23} & 31.87\scriptsize{±0.00} & 59.48\scriptsize{±0.00} & 58.45\scriptsize{±0.04} & 51.84\scriptsize{±0.04} \\
 & \multirow{-3}{*}{\textbf{12B}} &  \multicolumn{1}{r|}{\textit{-Image}} & 53.45\scriptsize{±0.19} & 49.79\scriptsize{±0.20} & 57.97\scriptsize{±0.15} & 11.34\scriptsize{±0.05} & 35.95\scriptsize{±0.56} & 66.58\scriptsize{±0.28} & 61.78\scriptsize{±0.01} & 54.80\scriptsize{±0.08} \\
 \cline{2-11}
 &  & Meme & 61.32\scriptsize{±0.06} & 62.32\scriptsize{±0.05} & 62.51\scriptsize{±0.09} & 52.99\scriptsize{±0.24} & 63.64\scriptsize{±0.16} & 62.82\scriptsize{±0.15} & 63.38\scriptsize{±0.01} & 68.21\scriptsize{±0.01} \\
 &  &  \multicolumn{1}{r|}{\textit{-Text}\phantom{ }\phantom{ }\phantom{ }} & 58.66\scriptsize{±0.04} & 60.87\scriptsize{±0.00} & 42.68\scriptsize{±0.26} & 53.81\scriptsize{±0.08} & 52.77\scriptsize{±0.12} & 61.65\scriptsize{±0.11} & 60.46\scriptsize{±0.04} & 53.99\scriptsize{±0.02} \\
\multirow{-9}{*}{\textbf{Gemma}} & \multirow{-3}{*}{\textbf{27B}} &  \multicolumn{1}{r|}{\textit{-Image}} & 54.66\scriptsize{±0.06} & 41.53\scriptsize{±0.09} & 60.68\scriptsize{±0.22} & 13.78\scriptsize{±0.06} & 47.18\scriptsize{±0.16} & 66.80\scriptsize{±0.27} & 62.77\scriptsize{±0.02} & 57.65\scriptsize{±0.07} \\
\midrule\midrule
 &  & Meme & 59.92\scriptsize{±0.08} & 52.96\scriptsize{±0.18} & 52.58\scriptsize{±0.22} & 48.52\scriptsize{±1.27} & 9.10\scriptsize{±0.41} & 75.62\scriptsize{±0.37} & 63.22\scriptsize{±0.00} & 70.90\scriptsize{±0.05} \\
 &  &  \multicolumn{1}{r|}{\textit{-Text}\phantom{ }\phantom{ }\phantom{ }}  & 56.77\scriptsize{±0.08} & 53.26\scriptsize{±0.18} & 9.94\scriptsize{±0.26} & 48.49\scriptsize{±0.44} & 3.27\scriptsize{±0.01} & 71.36\scriptsize{±0.20} & 36.60\scriptsize{±0.03} & 36.22\scriptsize{±0.00} \\
 & \multirow{-3}{*}{\textbf{7B}} &  \multicolumn{1}{r|}{\textit{-Image}} &  23.48\scriptsize{±0.00} & 30.33\scriptsize{±0.00} & 26.20\scriptsize{±0.24} & 1.77\scriptsize{±0.00} & 6.39\scriptsize{±0.01} & 45.95\scriptsize{±0.36} & 56.85\scriptsize{±0.00} & 39.29\scriptsize{±0.00} \\
 \cline{2-11}
 &  & Meme & 62.48\scriptsize{±0.15} & 60.06\scriptsize{±0.36} & 69.81\scriptsize{±0.14} & 56.99\scriptsize{±0.16} & 64.34\scriptsize{±0.25} & 68.67\scriptsize{±0.07} & 62.58\scriptsize{±0.02} & 69.48\scriptsize{±0.04} \\
 &  &  \multicolumn{1}{r|}{\textit{-Text}\phantom{ }\phantom{ }\phantom{ }}  & 64.78\scriptsize{±0.07} & 63.66\scriptsize{±0.10} & 13.56\scriptsize{±0.24} & 53.51\scriptsize{±0.00} & 31.80\scriptsize{±0.27} & 68.68\scriptsize{±0.13} & 27.00\scriptsize{±0.18} & 37.65\scriptsize{±0.02} \\
 & \multirow{-3}{*}{\textbf{32B}} &  \multicolumn{1}{r|}{\textit{-Image}} & 56.69\scriptsize{±0.00} & 38.21\scriptsize{±0.00} & 53.36\scriptsize{±0.00} & 17.78\scriptsize{±0.00} & 32.63\scriptsize{±0.00} & 65.56\scriptsize{±0.00} & 56.85\scriptsize{±0.09} & 57.85\scriptsize{±0.05} \\
 \cline{2-11}
 &  & Meme & 65.31\scriptsize{±0.17} & 61.61\scriptsize{±0.06} & 65.35\scriptsize{±0.04} & 54.09\scriptsize{±0.17} & 61.76\scriptsize{±0.19} & 76.76\scriptsize{±0.38} & 62.81\scriptsize{±0.00} & 70.57\scriptsize{±0.18} \\
 &  &  \multicolumn{1}{r|}{\textit{-Text}\phantom{ }\phantom{ }\phantom{ }}  & 66.18\scriptsize{±0.08} & 60.93\scriptsize{±0.09} & 22.95\scriptsize{±0.26} & 55.24\scriptsize{±0.33} & 41.63\scriptsize{±0.28} & 74.90\scriptsize{±0.19} & 42.90\scriptsize{±0.02} & 50.53\scriptsize{±0.40} \\
\multirow{-9}{*}{\textbf{Qwen}} & \multirow{-3}{*}{\textbf{72B}} &  \multicolumn{1}{r|}{\textit{-Image}} &  53.80\scriptsize{±0.00} & 37.96\scriptsize{±0.00} & 61.81\scriptsize{±0.00} & 16.92\scriptsize{±0.00} & 30.53\scriptsize{±0.00} & 68.46\scriptsize{±0.00} & 60.30\scriptsize{±0.01} & 55.62\scriptsize{±0.05} \\
\bottomrule
\end{tabular}
}
\caption{F1 scores (in \%) of eight models on the \textit{binary} classification of figurative meaning types in memes from three datasets. 
\textit{-Text} and \textit{-Image} denote the ablation of the text or image modality, respectively.
Following \citet{liu2022FigMemes}, in \figmemes{} we consider memes with 0 labels as negative cases, and take the best-performing benchmark score as baseline.
Results are averaged over five runs.
}
\label{tab:main_result_std}
\end{table*}

\begin{table*}[t]
\centering
\resizebox{0.98\linewidth}{!}{
\begin{tabular}{lll|*{6}{>{\raggedleft\arraybackslash}p{1.5cm}}|r|r}
\toprule
 &  &  & \multicolumn{6}{c|}{\textbf{\figmemes}} & \multicolumn{1}{r|}{\textbf{\memotion}} & \multicolumn{1}{r}{\textbf{\metmeme}} \\
\multirow{-2}{*}{\textbf{Model}} & 
\multirow{-2}{*}{\textbf{Size}} & 
\multirow{-2}{*}{\textbf{Input\phantom{ }\phantom{ }\phantom{ }\phantom{ }\phantom{ }\phantom{ }}} & 
\multicolumn{1}{c}{\textbf{Allusion}} & 
\multicolumn{1}{c}{\textbf{Exag.}} & 
\multicolumn{1}{c}{\textbf{Irony}} & 
\multicolumn{1}{c}{\textbf{Anthrop.}} & 
\multicolumn{1}{c}{\textbf{Metaphor}} & 
\multicolumn{1}{c|}{\textbf{Contrast}} & 
\multicolumn{1}{c|}{\textbf{Sarcasm}} & 
\multicolumn{1}{c}{\textbf{Metaphor}} \\
\midrule
\multicolumn{3}{c|}{\textit{\textbf{Baseline}}} & 52.32 & 44.00 & 49.77 & 41.76 & 44.87 & 56.91 & -- & -- \\\midrule
 &  & Meme & 60.97\scriptsize{±0.05} & 58.72\scriptsize{±0.11} & 59.99\scriptsize{±0.03} & 49.13\scriptsize{±0.07} & 53.19\scriptsize{±0.13} & 59.80\scriptsize{±0.15} & 63.27\scriptsize{±0.03} & 66.88\scriptsize{±0.01} \\
 &  &  \multicolumn{1}{r|}{\textit{-Text}\phantom{ }\phantom{ }\phantom{ }} & 61.41\scriptsize{±0.05} & 59.42\scriptsize{±0.08} & 46.37\scriptsize{±0.00} & 49.39\scriptsize{±0.05} & 53.68\scriptsize{±0.06} & 57.01\scriptsize{±0.15} & 59.17\scriptsize{±0.02} & 63.51\scriptsize{±0.04} \\
 & \multirow{-3}{*}{\textbf{8B}} &  \multicolumn{1}{r|}{\textit{-Image}} & 51.66\scriptsize{±0.15} & 30.95\scriptsize{±0.08} & 58.17\scriptsize{±0.06} & 6.43\scriptsize{±0.04} & 52.34\scriptsize{±0.13} & 45.69\scriptsize{±0.00} & 56.03\scriptsize{±0.04} & 53.62\scriptsize{±0.05} \\
 \cline{2-11}
 &  & Meme & 69.11\scriptsize{±0.00} & 57.43\scriptsize{±0.04} & 60.55\scriptsize{±0.00} & 54.56\scriptsize{±0.10} & 57.41\scriptsize{±0.05} & 53.47\scriptsize{±0.04} & 63.21\scriptsize{±0.00} & 67.25\scriptsize{±0.00} \\ 
 &  &  \multicolumn{1}{r|}{\textit{-Text}\phantom{ }\phantom{ }\phantom{ }} & 66.05\scriptsize{±0.00} & 57.29\scriptsize{±0.00} & 43.72\scriptsize{±0.04} & 54.49\scriptsize{±0.00} & 55.69\scriptsize{±0.28} & 48.31\scriptsize{±0.08} & 52.69\scriptsize{±0.05} & 63.12\scriptsize{±0.02} \\
\multirow{-6}{*}{\textbf{Aya}}  & \multirow{-3}{*}{\textbf{32B}} &  \multicolumn{1}{r|}{\textit{-Image}} & 49.40\scriptsize{±0.00} & 41.75\scriptsize{±0.05} & 61.93\scriptsize{±0.04} & 12.12\scriptsize{±0.00} & 46.79\scriptsize{±0.07} & 51.58\scriptsize{±0.16} & 59.24\scriptsize{±0.04} & 49.47\scriptsize{±0.11} \\
\midrule\midrule
 &  & Meme & 68.25\scriptsize{±0.57} & 52.29\scriptsize{±0.36} & 60.95\scriptsize{±0.47} & 45.66\scriptsize{±0.94} & 52.14\scriptsize{±0.16} & 43.75\scriptsize{±0.43} & 63.47\scriptsize{±0.02} & 66.75\scriptsize{±0.02} \\
 &  &  \multicolumn{1}{r|}{\textit{-Text}\phantom{ }\phantom{ }\phantom{ }}  &  60.52\scriptsize{±0.45} & 55.78\scriptsize{±0.26} & 45.17\scriptsize{±0.40} & 49.62\scriptsize{±0.36} & 53.34\scriptsize{±0.30} & 50.74\scriptsize{±0.33} & 61.30\scriptsize{±0.02} & 61.75\scriptsize{±0.01} \\
 & \multirow{-3}{*}{\textbf{4B}} &  \multicolumn{1}{r|}{\textit{-Image}} & 35.48\scriptsize{±0.00} & 46.85\scriptsize{±0.12} & 58.46\scriptsize{±0.04} & 6.52\scriptsize{±0.03} & 48.48\scriptsize{±0.04} & 53.52\scriptsize{±0.00} & 63.37\scriptsize{±0.00} & 60.31\scriptsize{±0.00} \\
 \cline{2-11}
 &  & Meme & 61.29\scriptsize{±0.00} & 58.72\scriptsize{±0.04} & 61.86\scriptsize{±0.00} & 52.51\scriptsize{±0.00} & 52.70\scriptsize{±0.05} & 55.54\scriptsize{±0.04} & 63.39\scriptsize{±0.00} & 64.91\scriptsize{±0.03} \\
 &  &  \multicolumn{1}{r|}{\textit{-Text}\phantom{ }\phantom{ }\phantom{ }}  &  63.42\scriptsize{±0.00} & 57.75\scriptsize{±0.00} & 46.98\scriptsize{±0.00} & 54.16\scriptsize{±0.09} & 41.09\scriptsize{±0.00} & 57.35\scriptsize{±0.00} & 54.51\scriptsize{±0.08} & 47.50\scriptsize{±0.12} \\
 & \multirow{-3}{*}{\textbf{12B}} &  \multicolumn{1}{r|}{\textit{-Image}} &  53.62\scriptsize{±0.00} & 45.45\scriptsize{±0.11} & 57.93\scriptsize{±0.05} & 6.78\scriptsize{±0.00} & 41.62\scriptsize{±0.34} & 69.55\scriptsize{±0.28} & 61.07\scriptsize{±0.00} & 54.41\scriptsize{±0.11} \\
 \cline{2-11}
 &  & Meme & 59.13\scriptsize{±0.52} & 60.75\scriptsize{±0.39} & 61.80\scriptsize{±0.17} & 52.04\scriptsize{±0.44} & 62.73\scriptsize{±0.44} & 64.16\scriptsize{±0.58} & 63.42\scriptsize{±0.05} & 68.67\scriptsize{±0.05} \\
 &  &  \multicolumn{1}{r|}{\textit{-Text}\phantom{ }\phantom{ }\phantom{ }} & 60.02\scriptsize{±0.04} & 61.73\scriptsize{±0.09} & 52.79\scriptsize{±0.27} & 51.90\scriptsize{±0.00} & 54.37\scriptsize{±0.28} & 59.43\scriptsize{±0.06} & 57.34\scriptsize{±0.05} & 57.90\scriptsize{±0.05} \\
\multirow{-9}{*}{\textbf{Gemma}} & \multirow{-3}{*}{\textbf{27B}} &  \multicolumn{1}{r|}{\textit{-Image}} & 54.82\scriptsize{±0.03} & 41.91\scriptsize{±0.00} & 59.95\scriptsize{±0.17} & 14.69\scriptsize{±0.05} & 49.44\scriptsize{±0.14} & 64.35\scriptsize{±0.14} & 60.38\scriptsize{±0.03} & 60.17\scriptsize{±0.03} \\
\midrule\midrule
 &  & Meme & 62.60\scriptsize{±0.00} & 51.23\scriptsize{±0.17} & 58.77\scriptsize{±0.15} & 51.18\scriptsize{±0.00} & 30.53\scriptsize{±0.26} & 61.20\scriptsize{±0.22} & 62.21\scriptsize{±0.03} & 72.26\scriptsize{±0.06} \\
 &  &  \multicolumn{1}{r|}{\textit{-Text}\phantom{ }\phantom{ }\phantom{ }} & 61.85\scriptsize{±0.08} & 53.27\scriptsize{±0.05} & 20.96\scriptsize{±0.27} & 47.93\scriptsize{±0.32} & 31.31\scriptsize{±0.17} & 57.71\scriptsize{±0.15} & 31.80\scriptsize{±0.17} & 27.43\scriptsize{±0.00} \\
 & \multirow{-3}{*}{\textbf{7B}} &  \multicolumn{1}{r|}{\textit{-Image}} & 24.63\scriptsize{±0.00} & 29.65\scriptsize{±0.00} & 40.82\scriptsize{±0.00} & 6.72\scriptsize{±0.00} & 30.05\scriptsize{±0.00} & 53.73\scriptsize{±0.06} & 44.17\scriptsize{±0.00} & 32.62\scriptsize{±0.00} \\
 \cline{2-11}
 &  & Meme & 61.76\scriptsize{±0.04} & 59.10\scriptsize{±0.07} & 69.79\scriptsize{±0.13} & 56.07\scriptsize{±0.09} & 63.50\scriptsize{±0.32} & 60.21\scriptsize{±0.29} & 62.37\scriptsize{±0.02} & 72.54\scriptsize{±0.06} \\
 &  &  \multicolumn{1}{r|}{\textit{-Text}\phantom{ }\phantom{ }\phantom{ }}  & 65.67\scriptsize{±0.10} & 60.07\scriptsize{±0.16} & 24.03\scriptsize{±0.61} & 52.72\scriptsize{±0.07} & 52.39\scriptsize{±0.64} & 62.74\scriptsize{±0.39} & 28.65\scriptsize{±0.08} & 28.12\scriptsize{±0.01} \\
 & \multirow{-3}{*}{\textbf{32B}} &  \multicolumn{1}{r|}{\textit{-Image}} & 54.60\scriptsize{±0.04} & 42.71\scriptsize{±0.00} & 54.97\scriptsize{±0.08} & 18.92\scriptsize{±0.00} & 46.50\scriptsize{±0.10} & 61.36\scriptsize{±0.08} & 56.31\scriptsize{±0.07} & 49.68\scriptsize{±0.06} \\
 \cline{2-11}
 &  & Meme & 67.61\scriptsize{±0.24} & 57.83\scriptsize{±0.15} & 63.95\scriptsize{±0.22} & 53.98\scriptsize{±0.28} & 63.99\scriptsize{±0.51} & 70.42\scriptsize{±0.53} & 62.68\scriptsize{±0.00} & 74.06\scriptsize{±0.01} \\
 &  &  \multicolumn{1}{r|}{\textit{-Text}\phantom{ }\phantom{ }\phantom{ }}  & 64.63\scriptsize{±0.07} & 60.07\scriptsize{±0.09} & 37.83\scriptsize{±0.34} & 55.14\scriptsize{±0.27} & 50.73\scriptsize{±0.36} & 67.59\scriptsize{±0.20} & 49.17\scriptsize{±0.02} & 47.74\scriptsize{±0.00} \\
\multirow{-9}{*}{\textbf{Qwen}} & \multirow{-3}{*}{\textbf{72B}} &  \multicolumn{1}{r|}{\textit{-Image}} & 53.61\scriptsize{±0.00} & 38.81\scriptsize{±0.00} & 60.76\scriptsize{±0.05} & 13.85\scriptsize{±0.00} & 40.52\scriptsize{±0.11} & 69.16\scriptsize{±0.00} & 58.86\scriptsize{±0.06} & 51.08\scriptsize{±0.05} \\
\bottomrule
\end{tabular}
}
\caption{F1 scores (in \%) of eight models on the \textit{binary} classification of figurative meaning types in memes from three datasets with the second prompt template. 
\textit{-Text} and \textit{-Image} denote the ablation of the text or image modality, respectively.
Following \citet{liu2022FigMemes}, in \figmemes{} we consider memes with 0 labels as negative cases, and take the best-performing benchmark score as baseline.
Results are averaged over five runs.
}
\label{tab:main_result_std_second}
\end{table*}

\end{document}